# 视频单目标跟踪研究进展综述

韩瑞泽[1),2)] 冯伟[1),2)] 郭青[1),2)] 胡清华[1)]

[1)]天津大学智能与计算学部, 天津 300350

[2)]文物本体表面监测与分析研究国家文物局重点科研基地, 天津 300350

**摘 要** 视频目标跟踪是计算机视觉中的重要任务之一，在实际生活中有着广泛的应用，例如视频监控、视觉导航等。视频目标跟踪任务也面临着诸多挑战，如目标遮挡、目标形变等情形。为解决目标跟踪中的挑战，实现精确高效的目标跟踪，近年来出现大量的目标跟踪算法。本文介绍了近十年来视频目标跟踪领域两大主流算法框架（基于相关滤波和孪生网络的目标跟踪算法）的基本原理、改进策略和代表性工作，之后按照网络结构分类介绍了其他基于深度学习的目标跟踪算法，还从解决目标跟踪所面临挑战的角度介绍了应对各类问题的典型解决方案，并总结了视频目标跟踪的历史发展脉络和未来发展趋势。本文还详细介绍和比较了面向目标跟踪任务的数据集和挑战赛，并从数据集的数据统计和算法的评估结果出发，总结了各类视频目标跟踪算法的特点和优势。针对目标跟踪未来发展趋势，本文认为视频目标跟踪还面临诸多难题亟需解决，例如当前的算法往往无法在长时间、低功耗、抗干扰的环境下实地应用。未来，考虑多模态数据融合，如将深度图像、红外图像数据与传统彩色视频联合分析，将会为目标跟踪带来更多新的解决方案。目标跟踪任务也将会和其他任务，如视频目标检测、视频目标分割，相互促进共同发展。

## Single Object Tracking Research : A Survey

HAN Rui-Ze[1)2)] FENG Wei[1)2)] GUO Qing[1)2)] HU Qing-Hua[1)]

[1)] College of Intelligence and Computing, Tianjin University, Tianjin 300350

[2)]Key Research Center for Surface Monitoring and Analysis of Cultural Relics, Tianjin 300350

**Abstract** Visual object tracking is an important and fundamental task in computer vision, which has many real-world applications, e.g., video surveillance, visual navigation and robotic service. Visual object tracking also has many challenges, such as object loss, object deformation, background clutters, and object fast motion. To solve the above problems and track the target accurately and efficiently, many visual object tracking algorithms have been emerged in recent years. In this paper, we first review the two most popular tracking frameworks in the past ten years, i.e., the Correlation Filter (CF) and Siamese network based visual object tracking. We present the rationale, the improvement strategy, and the representative works of the above two frameworks in detail. Specifically, the CF technology has been used in visual object tracking for over ten years, which has a good balance between the tracking accuracy and running speed. In CF tracking, the target is located by applying a circular convolution operation on the learned filter and the current frame, which can be efficiently achieved by

——————————————

收稿日期：年-月-日；最终修改稿收到日期：年-月-日 *投稿时不填写此项*. 本课题得到天津市自然科学基金项目 No. 18JCYBJC15200，天津市研究生科研创新项目 No. 2021YJSB174，国家自然科学基金项目 (NSFC) No. U1803264资助. 韩瑞泽，男，1994年生，博士生，计算机学会 (CCF) 会员 (D3683G)，主要研究领域为多相机视频分析、视频目标跟踪. E-mail: han_ruize@tju.edu.cn. 冯伟（通信作者），男，1978年生，博士，教授，计算机学会 (CCF) 会员 (C6681M)，主要研究领域为主动机器人视觉、视觉智能. E-mail: wfeng@tju.edu.cn. 郭青，男，1989年生，博士，主要研究领域为计算机视觉、目标跟踪. E-mail: tsingqguo@tju.edu.cn. 胡清华，男，1976年生，博士，计算机学会 (CCF) 会员，主要研究领域为大数据不确定性建模、多模态数据机器学习. E-mail: huqinghua@tju.edu.cn. 第1作者手机号码:15510905789, E-mail: han_ruize@tju.edu.cn.

the Fast Fourier Transform (FFT). The Siamese network based trackers locate the target from the candidate patches through a matching function offline learned on abundant training data in terms of image pairs. The matching function is modeled by a two-branch convolutional neural network (CNN) with shared parameters to learn the similarity between the target and the candidate patches. Besides the above two frameworks, we then present some other deep learning based tracking methods categorized by different network structures, e.g., RNN (Recurrent Neural Network), GCN (Graph Convolutional Network), etc. We also introduce some classical strategies for handling the challenges in the visual object tracking problem. From the recent tracking methods, we find that the development direction of the methods shows a diversified trend. More new network structures and skills have been applied to object tracking task. Although other deep tracking methods show a diversified trend, they have not formed a complete system. In the past ten years, the mainstream frameworks were the Correlation Filter and Siamese network. For the development trend of the visual tracking in the next few years, the CF tracking method has been relatively mature, and the development space in the future is limited. The deep learning algorithm based on CNN, especially the tracking algorithm under Siamese framework, could still be the mainstream framework. Further, this paper detailedly presents and compares the benchmarks and challenges for visual object tracking task, including the OTB benchmark, LaSOT benchmark, and VOT challenges, etc. Based on the data statistics of the datasets and the performance evaluation of the algorithms, we summarize the characteristics and advantages of various visual object tracking algorithms. For the future development of visual object tracking, which would be applied in real-world scenes before some problems to be addressed, such as the problems in long-term tracking, low-power high-speed tracking and attack-robust tracking. In the future, we can consider the integration of the traditional color (RGB) image together with the multi-modal data, such as the depth image, the thermal image, for joint analysis, which will provide more solutions for the visual object tracking task. Moreover, the visual tracking task will develop together with some other related tasks for mutual promotion, e.g., the video object detection, the video object segmentation task.

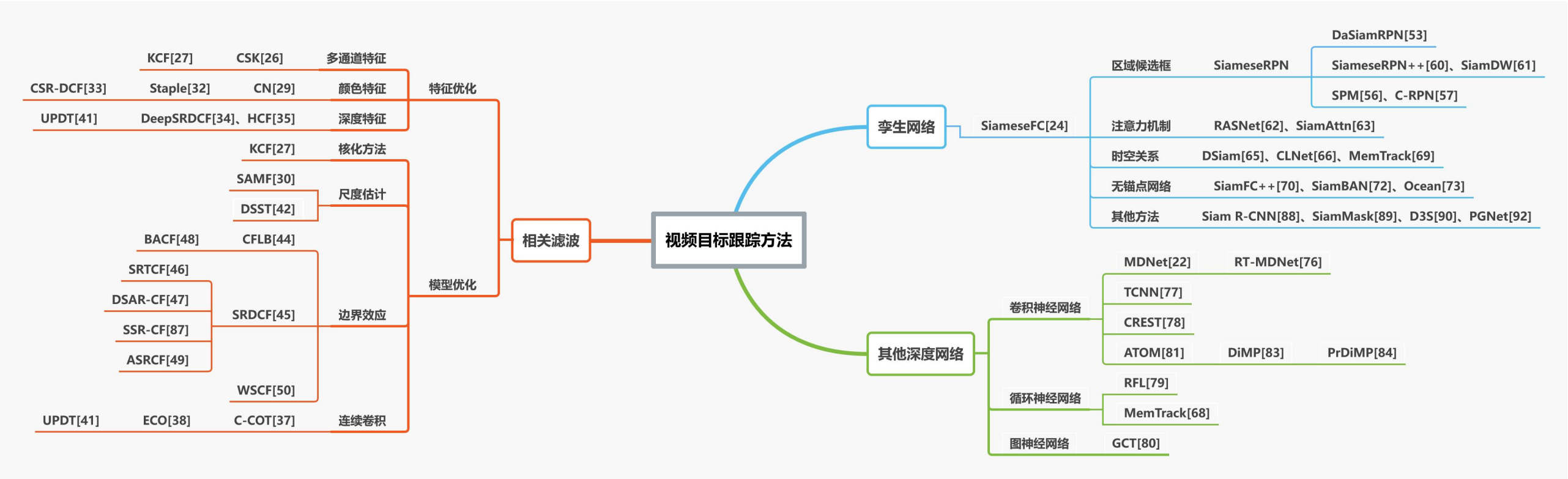

图 1 视频目标跟踪算法分类汇总

## 1 引言

视频目标跟踪是计算机视觉领域重要的基础性研究问题之一，是指在视频序列第一帧指定目标后，在后续帧持续跟踪目标，即利用边界框（通常用矩形框表示）标定目标，实现目标的定位与尺度估计（目标跟踪问题通常分为单目标跟踪和多目标跟踪，本文主要关注单目标跟踪问题）。视频目标跟踪具有广泛的应用价值，包括 1) 公共安防领域[1]：对人群或重点对象进行跟踪定位，实现监控场景下可疑人员轨迹重建与实时定位[2-4]；2) 自动驾驶领域[5]：辅助自主导航，轨迹规划等功能的实现；3) 智能机器人领域：用于机器人视觉导航，关注目

标的运动轨迹捕获与主动追踪；4) 人机智能交互领域：通过人体关键部位（如手部）跟踪与识别，实现计算机根据人体特定动作或手势等完成相应反馈。由于存在诸多技术挑战和潜在应用价值，视频目标跟踪近年来也引起学术界和工业界的广泛关注和大量研究[6,7]。视频目标跟踪的挑战主要体现在跟踪目标为非特定物体，且目标在视频序列往往会发生不可预期的变化和干扰。正是因为目标的非特定性，目标跟踪器无法预先对跟踪目标进行预先训练或建模。而在跟踪过程中，还会产生如目标消失、目标外观变化、背景干扰、目标快速移动等诸多问题，对目标跟踪造成严峻的挑战。

为解决目标跟踪问题中的困难，建立精确和高效的目标跟踪器，大量的目标跟踪算法应运而生。早期的目标跟踪采用了许多经典的机器学习方法，如支持向量机[8, 9]，集成学习[10]，稀疏重建[11]等。近年来，目标跟踪领域发展迅速。图 1 分类汇总了近十年来目标跟踪领域的主流算法框架和代表性方法。首先，2010 年，基于相关滤波的目标跟踪算法开始出现[12]，由于其具备良好的精度和超高的速度，迅速引起了相关研究者的广泛关注，围绕相关滤波算法框架，许多优化方法，如特征优化、模型优化应运而生，使得相关滤波目标跟踪算法发展成为近十年来目标跟踪的主流方法之一，在相关工作数量和各大数据集的性能表现上均具有明显优势。最近，深度学习在计算机视觉领域展现了强大的性能[13,14]，基于深度学习的目标跟踪算法也相继问世，其中孪生网络由于相 比于其他深度学习算法框架具备较高的计算速度，因此受到更广泛的关注和研究[15]，围绕孪生滤波网络的一系列方法也展现出强大的竞争力。另外，其他深度神经网络如卷积神经网络，循环神经网络以及图卷积神经网络也都在目标跟踪算法中得以应用，并展现出一定的优势。对于目标跟踪算法，本文首先以目标跟踪近年来的两大主流算法框架 — 相关滤波和孪生网络为主线，介绍两类方法的发展历程及具有代表性的相关工作，本文也将介绍其他深度学习框架下的相关算法。此外，本文还将重点介绍应对目标跟踪面临主要挑战问题的解决方案和代表性工作，包括上述提到的目标消失、目标外观变化、背景干扰、目标快速移等问题。

除了目标跟踪算法，算法评估数据集和挑战赛也是推动目标跟踪任务快速发展的重要动力之一。从最早期的 OTB[16] 数据集只包含 50 个视频，平均长度约 500 帧，到最新的 LaSOT[17] 数据集包含 1,400 个视频，平均长度超 2,500 帧。视频目标跟踪数据集正向大规模、长时间、多样化的方向一步步发展。本文也将详细介绍和比较近年来视频目标跟踪任务的数据集，包括 10 个普通彩色 (RGB) 视频数据集，1 个彩色-深度 (RGB-D) 视频以及 1 个彩色-红外 (RGB-T) 视频数据集。本文还介绍了目标跟踪主流挑战赛 VOT 的视频特点、评估方式等，以及近年来挑战赛的主要结果与分析。

尽管近年来目标跟踪算法在上述数据集上取得了较高的精度，但是视频目标跟踪距离实际应用还具有一定的差距。本文最后还从多个方面详尽讨论了目标跟踪未来的发展趋势。1）首先，针对目标跟踪发展面临的痛点，如目前算法无法适用于长时间、低功耗、抗干扰场景等，本文将重点介绍目标跟踪算法距离实际落地应用面临的瓶颈难题。例如，目前的目标跟踪数据集虽然视频长度已经较前些年明显增长，但是现实场景中往往需要实现分钟级别甚至小时级别的视频目标跟踪，因此实现长时间目标跟踪是未来的重要发展方向之一。此外，现有的跟踪算法尤其是基于深度网络或深度特征的算法，往往需要高性能设备支持，考虑到视频目标跟踪的应用场景，低功耗设备上的轻量级算法开发也是该领域的研究方向之一。面临深度学习需要大量训练数据的痛点，减少训练数据标注的弱监督、无监督方法也是目标跟踪未来的发展方向之一。还有考虑到算法的安全性及鲁棒性，针对目标跟踪算法的对抗攻击机制也开始兴起。此外，本文还涉及了特定场景，如无人机航拍视频，遥感图像下的目标跟踪研究。2）为实现更加鲁棒的跟踪，随着多模态数据采集设备的兴起与普及，考虑多视频源数据输入，如深度图像、红外图像等，用于视频目标跟踪，可以从数据源上有效地解决传统彩色视频中目标遮挡，光照变化等情形对跟踪带来的挑战。3）最后，为探究目标跟踪更多的应用长场景与交叉研究，本文还介绍了目标跟踪与计算机视觉领域其他密切相关任务，如视频目标检测、分割等问题的交叉研究。

本文后续章节的组织如下，第二章主要介绍视频目标跟踪任务面临的诸多挑战，第三章将分类介绍近十年来目标跟踪问题的主要方法，第四章介绍目标跟踪任务的主流评估数据集和挑战赛，以及相关算法在数据集和挑战赛上的评估结果和成绩，第

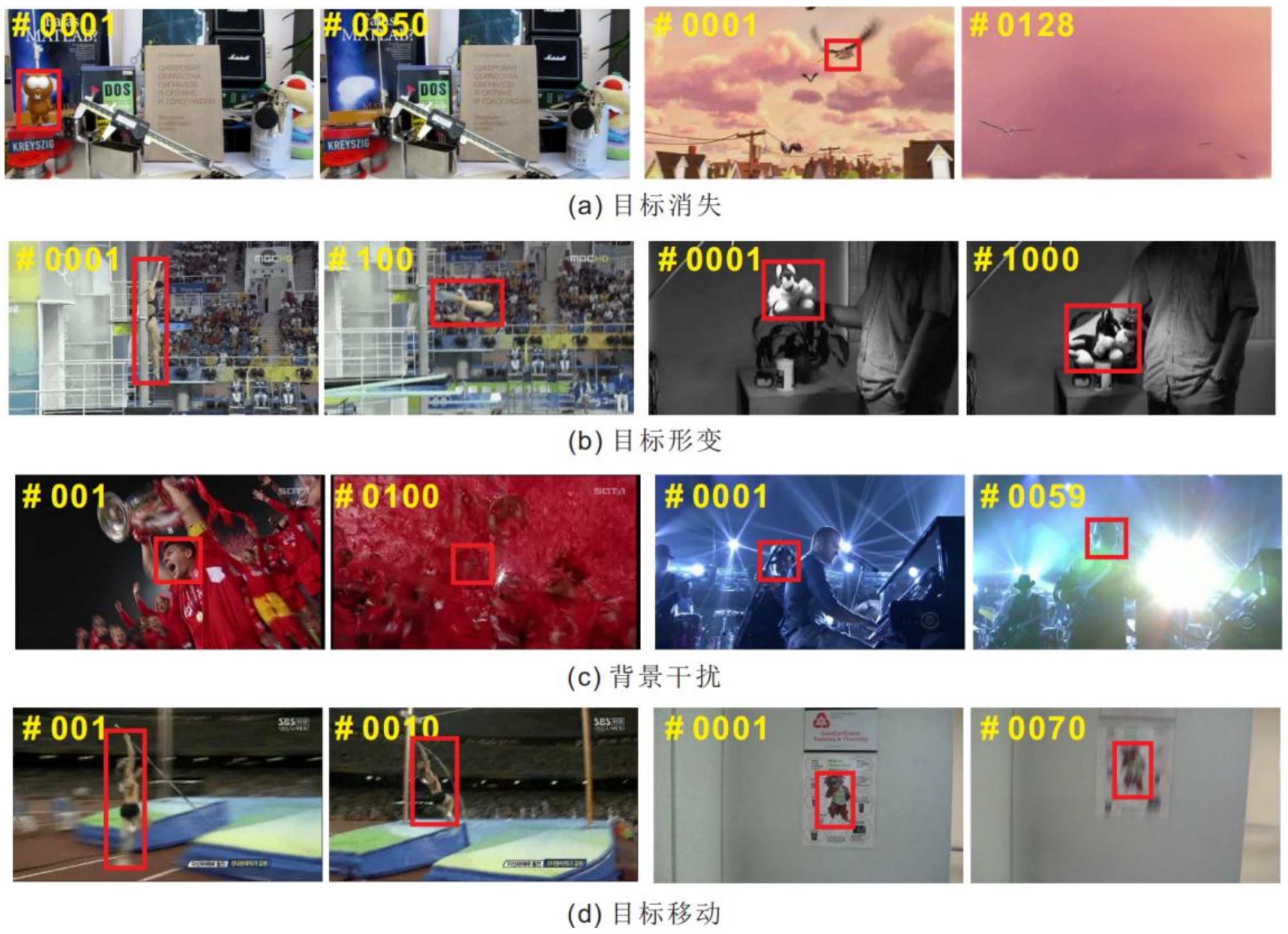

图 2 目标跟踪问题面临的主要挑战

五章展望了视频目标跟踪任务未来的发展趋势，最后，第六章对全文进行了总结。

## 2 视频目标跟踪中的挑战

对于视频目标跟踪问题，主要面临的挑战表现在视频目标前背景在跟踪过程中发生的复杂变化[18-20]。如图 2，这些变化包括：目标消失、目标形变、背景干扰以及目标移动等情形。上述情况往往导致视频序列中跟踪目标所依赖的特征，如外观、形状或背景等信息，随时间变化存在较大的不一致性，使得跟踪器在后续视频帧中无法准确识别和跟踪目标。

(1) 目标消失：目标消失是视频目标跟踪中最具挑战性的问题之一，主要包括在某段时间内目标（或部分目标）被其他物体遮挡或移出相机视野范围，当目标重新出现时如何继续跟踪目标，如图 2-(a) 所示。影响此类问题的因素主要包括遮挡范围和遮挡时间，若目标全部被遮挡或长时间被遮挡，往往会造成跟踪器无法有效更新，从而跟踪失败。

(2) 目标变化：目标变化是视频目标跟踪中最常见的问题之一，主要包括目标形变，目标旋转等情形。通常来说，非刚性物体在跟踪过程中都会发生不同程度的形变。如图 2-(b) 所示，左侧图示视频中目标（运动员）在执行动作过程中随时间发生了严重的形变，长宽比例变化明显。目标旋转通常包含两方面内容，一是平面内旋转，另一类是平面外旋转。前者是指目标旋转轴垂直于目标图像所在的平面，后者则表示旋转轴与图像平面不垂直的情形，图 2-(b) 右图展示了目标平面外旋转的例子。

(3) 背景干扰：背景干扰也是目标跟踪问题经常出现的问题，主要表现是背景杂乱和光照变化等情形。图 2-(c) 分别展示了目标受杂乱背景干扰和光照严重变化的情形。如何有效地进行前背景分离，从而精确地抓取前景抑制背景也是目标跟踪的根本问题。而光照变化不仅对背景造成干扰，也使得目标前景本身的外观特征发生一定程度的变化。强烈的光照变化通常造成不同帧序列之间目标外观差异增大，而同一帧之内目标前背景差异减小，从而加大跟踪的难度。

(4) 目标移动：视频目标跟踪所研究的对象主体往往是运动的目标，目标移动对目标跟踪造成的困难主要包括目标快速运动和目标运动模糊等情形。由于目标跟踪通常采取在目标前一帧所处位置周围区域进行搜索的策略，因此目标快速运动可能造成目标与前序帧位置差异较大，甚至超出搜索区域。另一方面，目标移动本身造成的运动模糊也会造成目标前景虚化，从而影响目标特征表达。同样的，相机移动甚至会造成整幅图像的模糊，也是影响目标跟踪效果的挑战之一。

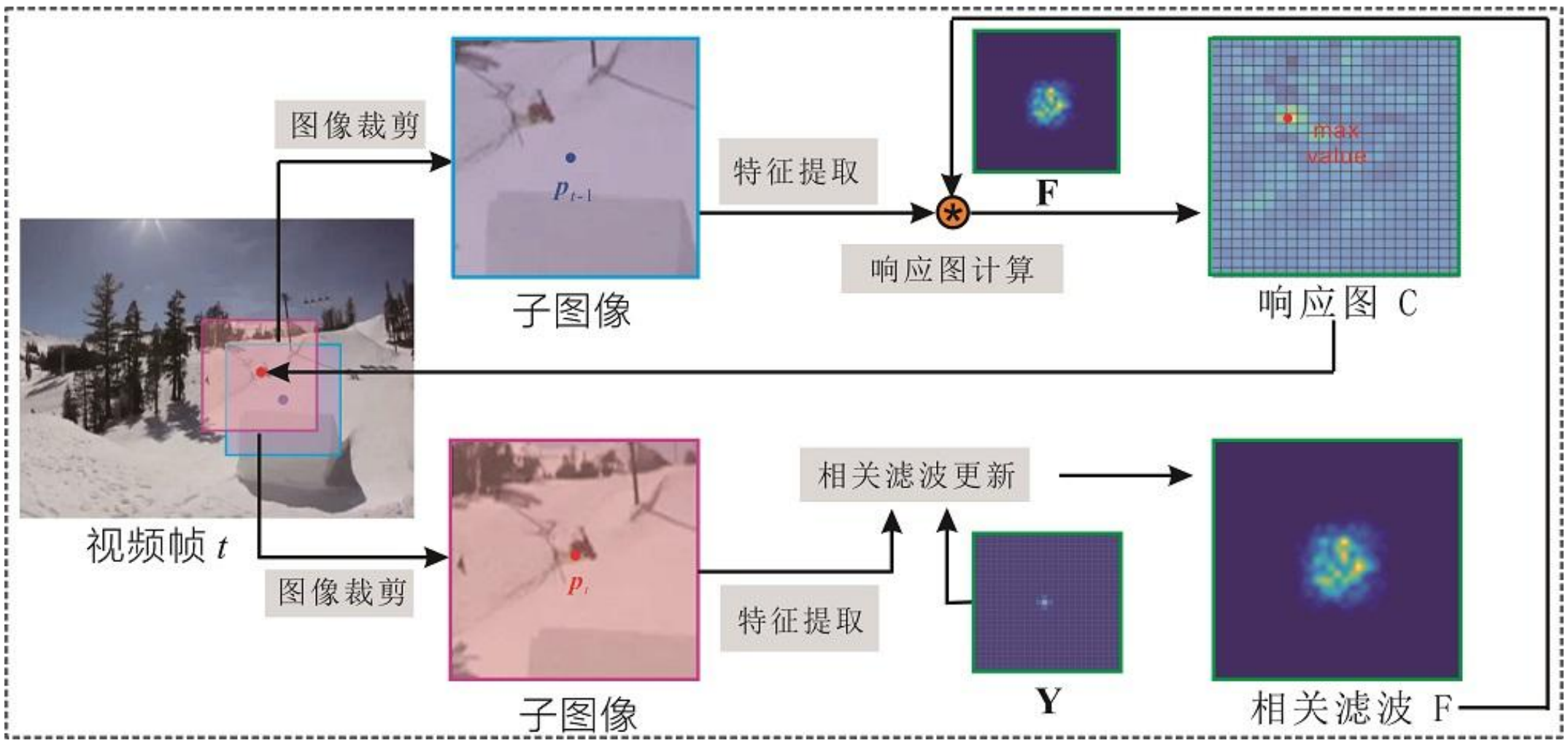

图 3 基于相关滤波的目标跟踪算法流程

最后，在实际的目标跟踪问题中，上述挑战往往并不是单一出现，多类困难并存的情形也十分常见。比如图 2-(b) 左侧示例中目标变形的同时也在快速移动，同样，图 2-(c) 左侧示例中目标背景干扰的同时也存在着形变和旋转。因此，尽管目标跟踪已经经历长时间的研究，其仍然存在着诸多挑战亟需解决。

## 3 视频目标跟踪方法

视频目标跟踪是计算机视觉的基础问题之一，近年来受到广泛关注和深入研究。许多机器学习的方法被广泛用于视频目标跟踪问题，例如: 支持向量机 SVM (Support Vector Machines)[8, 9]，增量学习[21]，集成学习[10]，稀疏重建[11]，相关滤波[12]，卷积神经网络 CNN (Convolutional Nueral Network)[22]，循环神经网络 RNN (Recurrent Neural Network)[23]等。其中，近十年来最主流的两类方法是基于相关滤波 CF (Correlation Filter) 和孪生网络 (Siamese Network) 框架的方法。相关滤波目标跟踪算法自 2010 年提出之后，由于其在跟踪精度和算法速度取得良好的平衡性，迅速发展成为目标跟踪的主流方法之一。自 2014 年以来，在视频目标跟踪主流挑战赛 VOT (Visula Object Tracking Challenge) 上，相关滤波目标跟踪算法在参赛数量和成绩上都具有明显的优势。基于孪生网络的目标跟踪算法相比相关滤波方法出现较晚，开创性工作是 2016 年出现的 SiameseFC[24] 算法。此后，基于孪生滤波的目标跟踪方法迅速发展，在文献数量和算法性能方面都取得显著的优势，在 VOT 挑战赛上也展现出一定的竞争力。本章将在第 3.1 节和 3.2 节分别介绍当前视频目标跟踪的两大主流方法，即基于相关滤波和孪生网络的目标跟踪，在第 3.3 节介绍其他基于深度神经网络的目标跟踪算法，第 3.4 节介绍解决目标跟踪所面临挑战而提出的方法和应对策略。

### 3.1 基于相关滤波的目标跟踪方法

基于相关滤波（CF）理论的目标跟踪算法在目标跟踪任务上取得了十分显著的进展，在跟踪精度及运行速度上具备明显的优势，是近年来视频目标跟踪任务的主流框架之一。CF 在计算机视觉的应用可以追溯至 20 世纪 80 年代，其最早应用于目标检测领域。 2010 年 MOSSE 算法[12] 第一次将 CF 应用于视频目标跟踪任务，在当时取得了出色的精度并具备超高的速度。相关滤波理论用于目标跟踪问题表现良好主要得益于以下两方面原因：1）CF 目标跟踪方法隐式地利用了循环平移操作对训练样本进行扩增，从而极大丰富了训练样本的多样性，使得算法的鲁棒性和精度提升；2）快速傅利叶变换 FFT (Fast Fourier Transform) 使得复杂的卷积操作在频域内加速计算，计算量降低，模型求解效率增加。

由于具备良好的精度以及速度，基于相关滤波理论的目标跟踪算法出现了大规模的发展，大量的相关滤波跟踪算法在公开数据集上展现了良好的效果。相关方法主要可以分为两类主要方向：1）利用更加强有力的特征提取方法；2）构建更加鲁棒的滤波器学习模型。接下来，本文将在小节 3.1.1 介绍相关滤波目标跟踪的基本思想和算法框架，在

小节 3.1.2 和 3.1.3 分别介绍使用更强力的特征和更鲁棒的模型，用于改进相关滤波目标跟踪的经典方法。

3.1.1 相关滤波目标跟踪框架

目标跟踪算法的输入是一段连续的视频序列，以及视频第一帧指定的跟踪目标（以矩形标定框 $B_1$ 的形式给出），目标跟踪算法的输出是在后续视频 $t > 1$ 中估计目标的位置以及大小，同样以标定框 $B_t$ 的形式给出。相关滤波目标跟踪算法的主要思想是，在当前帧更新相关滤波器（记作 $\boldsymbol{F}$），在下一帧利用所得的 $\boldsymbol{F}$ 通过循环卷积的操作实现目标中心点定位。不失一般性，我们考虑相关滤波目标跟踪算法在 $t-1$ 到第 $t$ 帧的算法流程。如图 3 所示，相关滤波视频目标跟踪算法主要包含以下 5 个步骤[25]：

*步骤 1*（搜索区域）：由于相邻两帧目标移动范围有限，利用第 $t-1$ 帧的跟踪结果 $B_{t-1}$，通过适当扩大 $B_{t-1}$ 得到目标搜索区域，并在视频第 $t$ 帧图像的上述搜索区域内进行目标定位搜索。

*步骤 2*（特征提取）：用步骤 1 得到的第 $t$ 帧的搜索区域，对该区域内的图像进行特征提取，得到特征图 $\boldsymbol{H}$。

*步骤 3*（目标定位）：相关滤波器 $\boldsymbol{F}$ 作用于提取的特征图 $\boldsymbol{H}$，利用公式 (1) 得到响应图 $\boldsymbol{C}$

$$\boldsymbol{C} = \boldsymbol{H} * \boldsymbol{F}, \tag{1}$$

式中 $*$ 为循环卷积，计算响应图 $\boldsymbol{C}$ 的最大值所在位置的坐标，即可得到当前帧图像上的目标中心位置，$B_t$ 的大小可由 $B_t - 1$ 进行缩放得到。

*步骤 4*（滤波更新）：利用当前跟踪结果，如图 3 下半栏所示，以目标为中心点截取子图像，类似步骤 2 提取特征图 $\boldsymbol{X}$，然后通过最小化公式 (2)，求解相关滤波器 $\boldsymbol{F}$

$$\mathrm{E}(\boldsymbol{F}) = \|\boldsymbol{X} * \boldsymbol{F} - \boldsymbol{Y}\|^2 + \|\boldsymbol{F}\|^2, \tag{2}$$

这里 $\boldsymbol{Y}$ 是以空间中心点为最高值的 2-D 高斯分布图。上述优化问题可以利用快速傅里叶变换（FFT）方法得到闭合解。

*步骤 5*（交替迭代）：令 $t = t + 1$，返回步骤 1 进行交替迭代。在视频每一帧重复上述步骤，可以逐帧得到滤波器以及每帧目标的位置及尺寸，完成视频目标跟踪任务。

在相关滤波目标跟踪算法基本框架下，大量的相关研究进一步开展，主要可以分为对特征和模型的优化两类。

3.1.2 特征优化

在最早的相关滤波目标跟踪算法 MOSSE[12] 中，作者仅采用了单通道灰度特征，就达到了与当时算法接近的精度，其速度更是达到了 600 fps 以上（单 CPU），在目标跟踪领域引起广泛关注，大量后续工作试图通过改进目标特征表示对算法进行进一步优化。

(1) 手工特征

对特征的改进首先体现在由采用单通道特征改变为融合多通道特征。Henriques 等人先后提出的著名的 CSK (Circulant Structure with Kernels) 和 KCF (Kernelized Correlation Filter) 算法[26, 27]是相关滤波目标跟踪算法中具有里程碑意义的工作。CSK 在 MOSSE 的基础上扩展了密集采样并采取了核化相关滤波方法，KCF 在 CSK 的基础上又进一步采用了多通道梯度特征 HOG (Histogram of Oriented Gradient) [28]。这使得基于相关滤波的目标跟踪算法超越了之前最优的方法，并仍然保持着超高的运行速度，CSK 的 CPU 运行速度超过 300 fps，KCF 也保持在 200 fps 以上。

Danelljan 等人[29] 最早考虑了颜色特征在视频目标跟踪的作用，综合评估了 RGB，LAB，YCbCr，rg，HSV 等各类颜色空间提取的特征在目标跟踪中的效果，并提出了基于多通道颜色特征 CN (Color Names)，这也成为后续相关滤波目标跟踪采用的主要手工颜色特征。在多通道特征应用之后，特征改进的下一个重要发展是融合不同类别的特征。Li 等人[30] 采用 HOG 特征和 CN 特征组合的方式，使得梯度特征 HOG 和颜色特征 CN 达到互补的效果，这也成为后续相关滤波目标跟踪算法最常用的手工特征。

Possegger 等人[31] 在 DAT (distractor-aware tracker) 方法中提出了颜色直方图特征，即统计前景目标和背景区域的颜色直方图并归一化，得到前背景颜色概率模型。将颜色直方图特征用于目标跟踪任务，通过逐像素判断其属于前景的概率，得到像素级前背景概率分布图，从而抑制与前景相似的干扰区域并缓解模型漂移 (model drift)。而后，Staple (Sum of Template And Pixel-wise LEarners) [32] 算法分析发现 HOG 对目标形变效果不好，但对光照变化等情况鲁棒；而 DAT 中的颜色直方图特征对变形不敏感，但对光照变化和前背景颜色相似敏感，于是将 HOG 特征和颜色直方图特征结合，得到了效果鲁棒且超实时（80 fps）的跟踪器 Staple，

算法效果与利用深度特征的方法接近，而且速度具有明显优势。另一结合颜色直方图特征的相关滤波方法是的 CSR-DCF (Discriminative Correlation Filter with Channel and Spatial Reliability) [33]，算法提出空间可靠性和通道可靠性的思想，计算目标前景空间分布和多特征通道权重分布，算法在 VOT 挑战赛上取得了良好的效果。近年来，随着深度学习的快速发展，基于深度神经网络的深度特征开始在相关滤波目标跟踪算法中应用普及。

(2) 深度特征

最近，由于深度卷积特征的广泛应用和优越效果，将深度特征集成到相关滤波目标跟踪算法也逐渐成为主流[34, 35]。DeepSRDCF[34] 将 HOG 特征替换为 VGG (Visual Geometry Group) 网络中单层卷积层的深度特征，使得目标跟踪精度相较于 HOG 特征有了很大提升。同期的工作 HCF (Hierarchical Convolutional Features for Visual Tracking)[35]，结合多层卷积特征（VGG 19 网络的 Conv5-4 , Conv4-4 和 Conv3-4 层）提升目标跟踪精度。

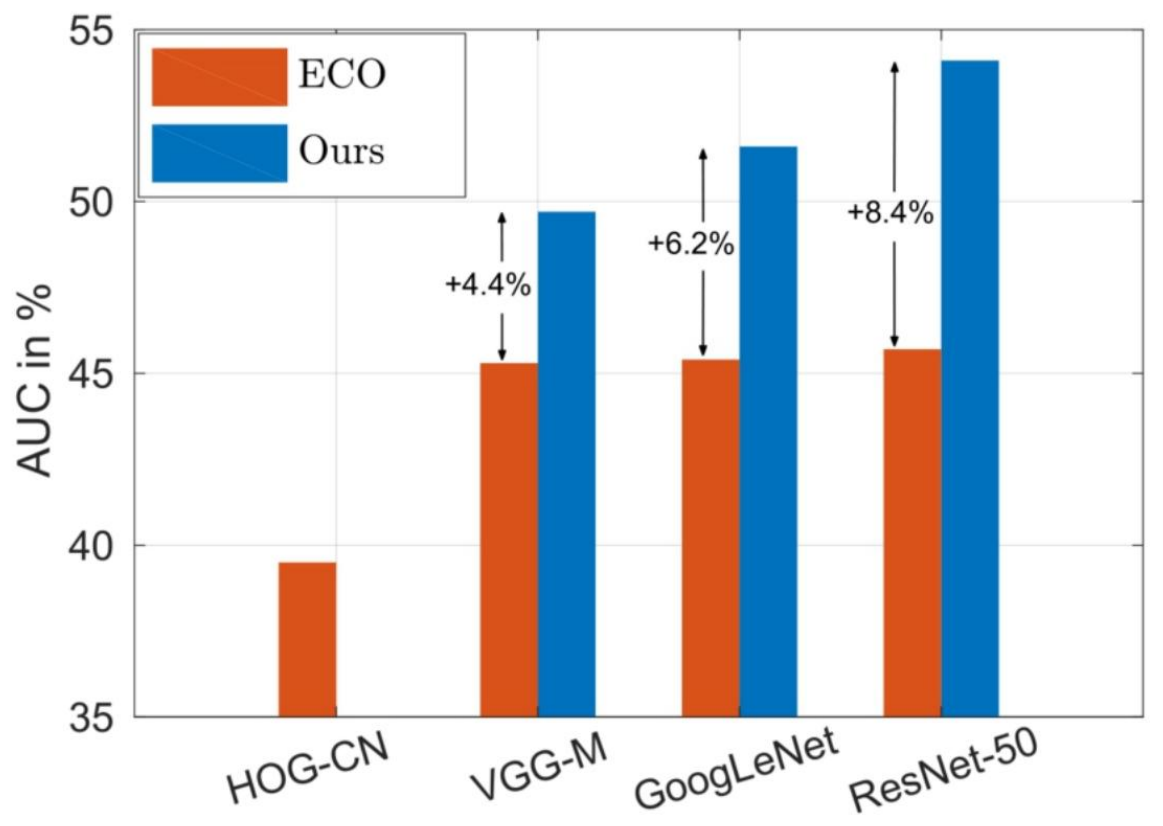

图 4. UPDT 算法利用不同深度特征的结果比较

深度特征展现出强大的性能表现后，大量的相关滤波目标跟踪方法[36-41] 开始采用深度特征。有研究尝试如果简单地将 VGG 网络换成更先进的 GoogLeNet 或 ResNet 提取深度特征，并没有像其他领域的研究一样对跟踪性能带来进一步的提升，如图 4 所示，经典目标跟踪算法 ECO (3.1.3 节介绍) 用更深的网络进行特征提取，跟踪性能并没有明显提升，这表明相关滤波目标跟踪方法无法从更深的卷积网络特征中获益。UPDT (Unveiling the Power of Deep Tracking)[41] 试图从这一问题入手，进一步发掘深度特征在目标跟踪中的潜力。文章发现深度特征可以更有效地表示高层语义信息，对目标旋转、变形等外观变化建模具备更强的鲁棒性；但同时平移和尺度不变性使得其无法精确定位目标，即准确性很差。相反，浅层特征（手工特征）可以更好地表示和建模纹理和颜色信息，保留高空间分辨率，更加适合准确的像素级目标定位；但是对旋转变形的鲁棒性很差。于是 UPDT 算法利用深度特征保持鲁棒性，同时采用浅层特征负责准确性，利用两种特征检测得到的响应图进行自适应融合，兼顾目标定位的准确性和跟踪的鲁棒性，从图 4 可以看到 UPDT 算法在 ECO 的基础上更好地发掘了深度特征的潜力和优势。

3.1.3 模型优化

(1) 核化方法

很多方法在相关滤波器的学习过程进行优化，其中最早的经典算法 CSK 和 KCF[26, 27] 是典型的代表。早期的 MOSSE[12] 提出了相关滤波理论用于视频目标跟踪问题的思想，而 CSK 和 KCF 正式提出了较成熟的相关滤波目标跟踪算法框架。作者利用了岭回归的核心思想，采用了循环移位对样本进行密集采样，并利用循环矩阵性质，缓和了目标样本密集采样带来的计算量问题，并推导了回归方法与相关滤波算法的等价性。作者同时提出了非线性核化方法将低维线性空间映射到高维空间，并给出了闭合解的形式，对相关滤波目标跟踪算法优化并提升算法精度和鲁棒性。

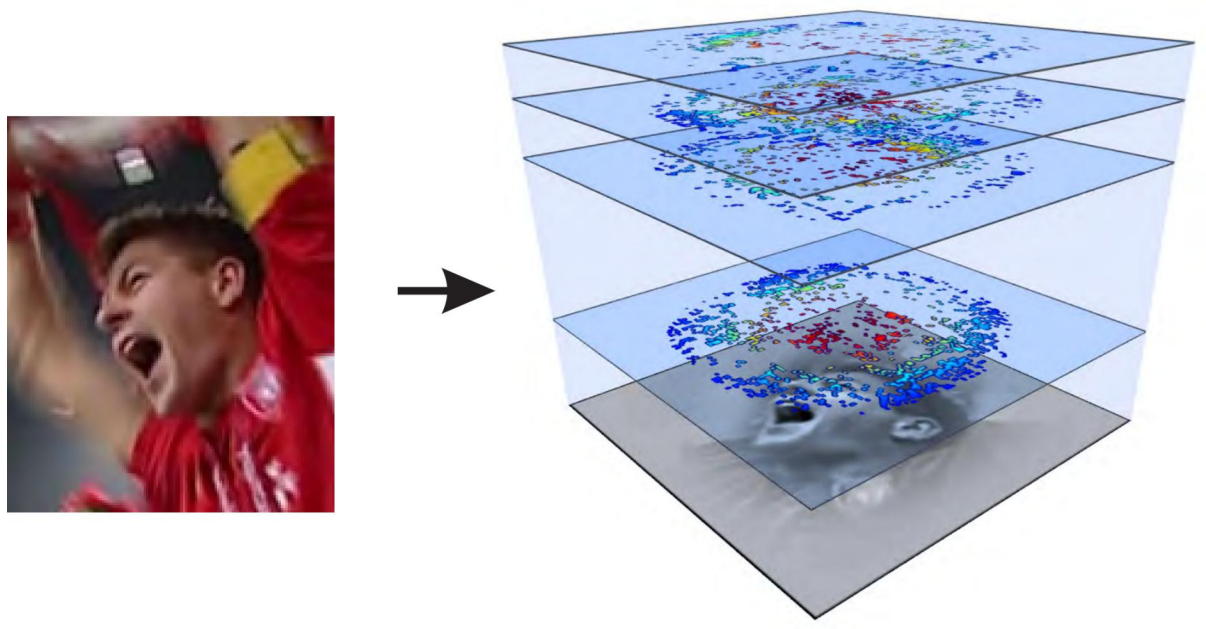
图 5. DSST 算法实现尺度估计

(2) 尺度估计方法

尺度估计是相关滤波目标跟踪算法模型优化的重要组成。在跟踪过程中，目标尺度大小经常发生变化，因此目标跟踪任务除了定位目标中心位置外，还需要估计其尺度大小。Danelljan 等人提出的 DSST (Discriminative Scale Space Tracking)[42] 算法首先利用基于 HOG 特征的相关滤波学习定位目标相邻帧的位置平移量，而后又训练独立的相关滤波器用于检测尺度变化，如图 5 所示，DSST 开创了平移检测滤波和尺度检测滤波相结合的方法。在

此基础上进一步改进的 fDSST (fast DSST)[43] 方法对 DSST 进行了加速优化，得到了更高效率的尺度检测方法。Li 等人提出的 SAMF[30] (Scale Adaptive with Multiple Features) 是另一类相关滤波目标跟踪的尺度检测方法。SAMF 直接对待检测区域进行固定次数的多尺度采样，并利用学得的滤波器在各个尺度进行目标检测，直接取最大响应值所在尺度作为估计的结果。相较而言，DSST 采用了 33 个不同大小的缩放尺度，而 SAMF 仅采用了 7 个，因此 DSST 尺度估计相对精确。而 SAMF 只需要学习一个滤波器，且可以同时得到目标跟踪任务的位置平移和尺度估计的最优解。在后续的工作中，由于简单而有效，SAMF 也得到了更广泛的应用，成为基于相关滤波目标跟踪尺度估计的重要手段。

(3) 边界效应问题

相关滤波目标跟踪算法中样本循环平移的思想在大大解决了训练样本匮乏问题的同时，边界效应 (boundary effects)[44] 的引入使得跟踪效果受到了一定程度的影响。边界效应是指循环平移产生的训练样本是合成样本，如图 6-(a) 所示，在样本生成过程中，目标边界随之循环平移，产生大量非真实训练样本，降低了模型的判别能力。

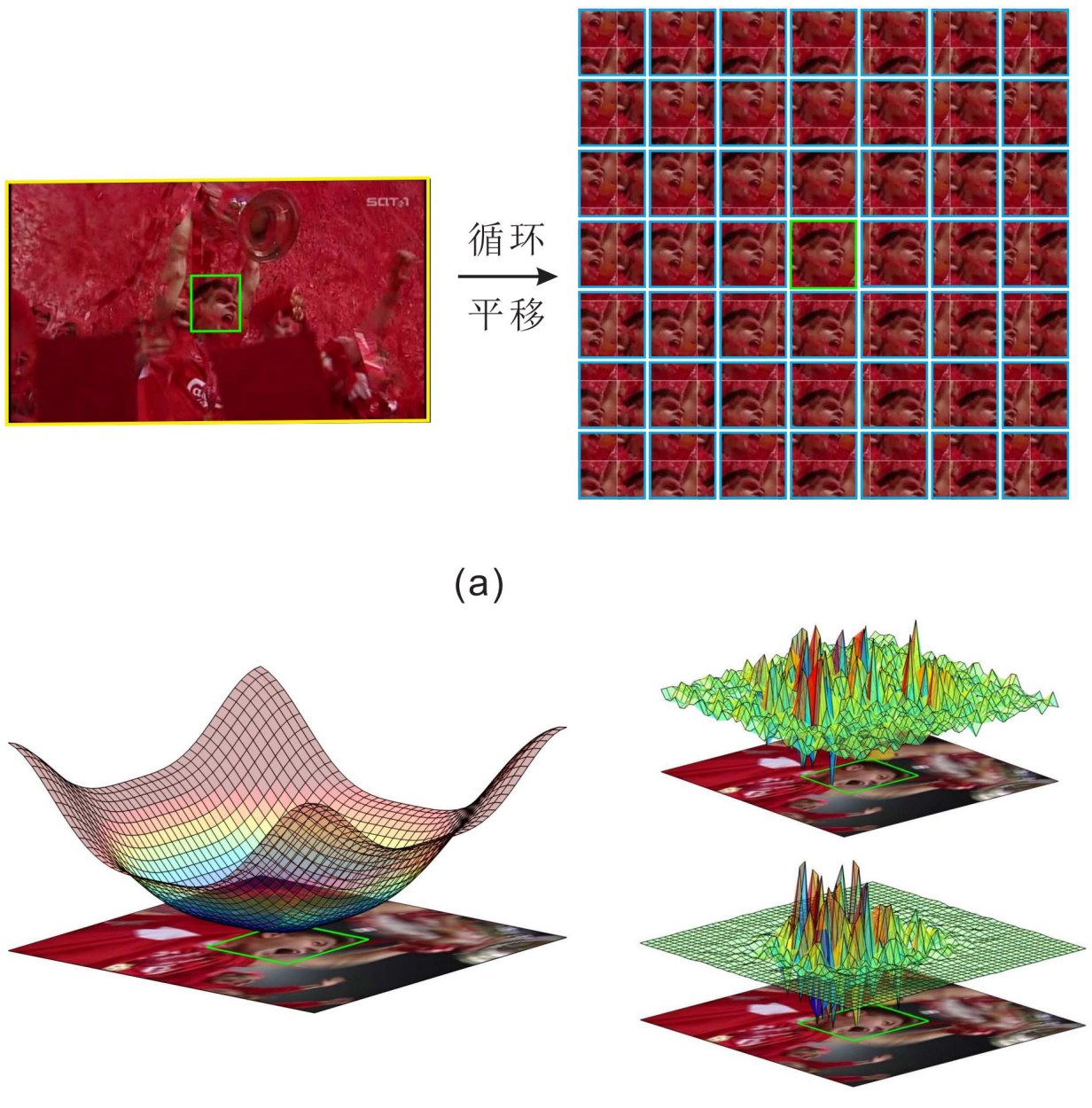

(a)

(b)

图 6. (a) 相关滤波目标跟踪算法循环平移生成样本

(b) SRDCF 算法空间正则化相关滤波

为解决边界效应的问题，经典方法之一是 SRDCF[45] 算法，其核心想法是考虑在相关滤波目标跟踪的目标函数中加入空间正则化项进行约束。具体来说，针对上述公式（2），在第 $t$ 帧更新滤波器 $\boldsymbol{F}$ 的过程中，引入空间权重图 $\boldsymbol{W}$ 作为约束项，得到新的优化目标函数

$$\mathrm{E}_{\mathrm{SR}}(\boldsymbol{F}) = \|\boldsymbol{X} * \boldsymbol{F} - \boldsymbol{Y}\|^2 + \|\boldsymbol{W} \odot \boldsymbol{F}\|^2, \quad (3)$$

式中第二项为正则化项，其中 $\odot$ 指矩阵元素乘操作，$\boldsymbol{W}$ 表示服从二次函数分布的空间权重图

$$\boldsymbol{W}(i,j) = a + \frac{b}{W^2}\left(i - \frac{M}{2}\right)^2 + \frac{b}{H^2}\left(j - \frac{N}{2}\right)^2, \quad (4)$$

如图 6-(b) 所示，其中 $i = 1, 2, \ldots, M$，$j = 1, 2, \ldots, N$。$a, b > 0$ 是设定的参数，$W, H$ 分别表示目标的宽和高。在相关滤波算法滤波器更新过程中中，由于目标中心处于采样图像的中心位置，所以正则化权重 $\boldsymbol{W}$ 的引入可以起到对滤波器 $\boldsymbol{F}$ 中非目标区域数值的抑制作用，缓解了边界效应问题。但是，由于 $\boldsymbol{W}$ 的引入，相关滤波算法在频域上的闭合解形式被破坏，使得公式 (3) 需要利用迭代法进行求解。因此， SRDCF 算法大大降低了相关滤波算法的计算速度，使得其无法满足目标跟踪的实时性需求。

SRDCF 作为目标跟踪经典方法，着眼考虑并有效缓解相关滤波跟踪算法的边界效应问题，是该体系下极具代表性的工作。后续很多工作在 SRDCF 的基础上开展，例如考虑到空间正则化方法的时序性，STRCF (Spatial-Temporal Regularized Correlation Filter)[46] 提出了时空域上的正则化方法，并采取 ADMM (Alternating Direction Method of Multipliers) 算法进行求解，使得 STRCF 达到了 CPU 上接近实时的运行速度。类似的，DSARCF (Dynamic Saliency-Aware Regularization Correlation Filter)[25, 47] 针对目标时序变化的问题，利用动态更新的显著图指导相关滤波空间正则化并实现其时序自适应调整。

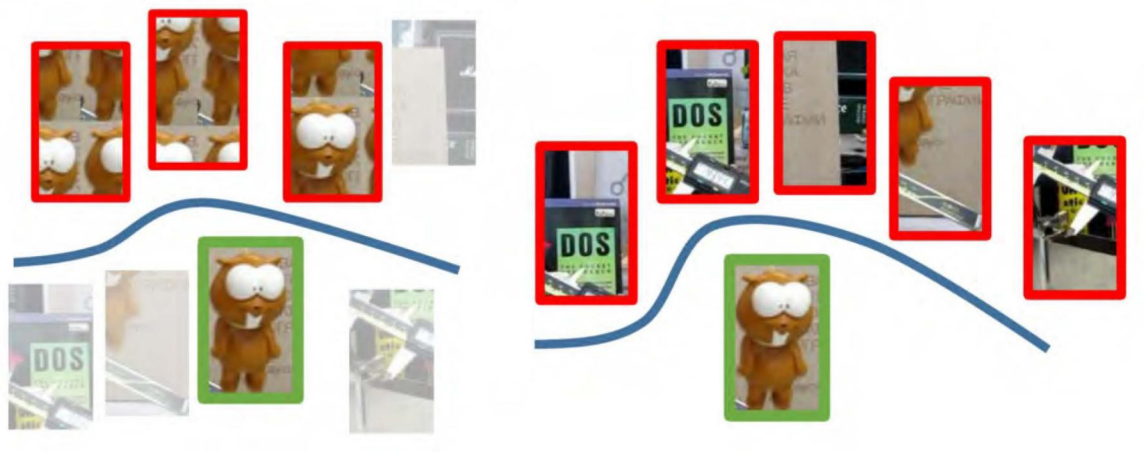

(a) (b)

图 7. BACF 算法利用背景信息生成更真实训练样本

另一类典型的缓解边界效应问题的方法是边界约束相关滤波方法 CFLB (Correlation filters with limited boundaries) 和背景指导相关滤波方法

BACF (Background-Aware Correlation Filters) [44, 48] 系列，见图 7。此类方法利用包含更大背景的区域用于目标检测和滤波器学习，与 SRDCF 的滤波器系数从中心到边缘连续过渡不同，CFLB 和 BACF 直接通过裁剪矩阵对滤波器边缘进行补零操作，得到作用域较小的相关滤波器。CFLB 与 MOSSE 类似仅采用单通道灰度特征，虽然速度高达 160 fps，但精度较差，而改进后的 BACF 将特征扩展为多通道 HOG 特征，并采取 ADMM 迭代法进行求解，不仅精度超过了 SRDCF，而且达到了实时的运行速度。在 SRDCF[45] 和 BACF[48] 的基础，ASRCF (Adaptive Spatially-Regularized Correlation Filter)[49] 兼顾了两者的思想和优势，与 DSARCF[47] 类似，提出了自适应空间正则化算法用于相关滤波目标跟踪，如图 8 所示，ASRCF 采用的空间正则化权重图在跟踪过程中可根据目标的特点进行自适应动态调整。同时，作者推导出 SRDCF 和 BACF 均可通过超参数设置由 ASRCF 变形得到，因此该方法具有更强的泛化能力，另外也同时兼顾了算法精度和速度，可以看做是基于空间正则化方法的相关滤波目标跟踪算法的集大成之作。

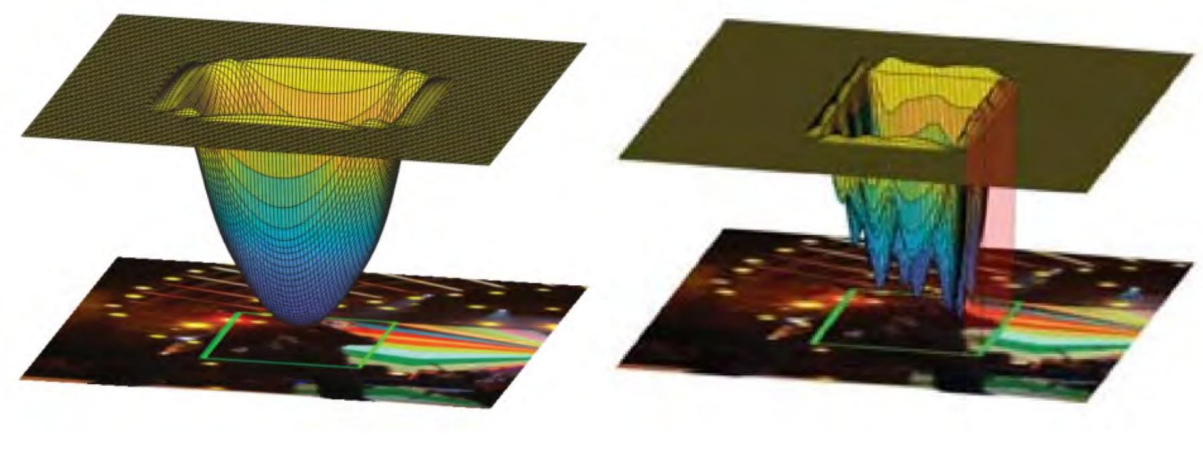

图 8. ASRCF 算法实现自适应空间正则化

与上述方法不同，WSCF (Weighted Sample based Correlation Filter)[50] 采用空间样本赋权重的思想解决边界效应问题，方法使循环平移严重的合成样本在训练过程中获得的权重较小，而接近真实条件下的训练样本权重增加。另外，由于该方法没有引入 SRDCF 中的正则化图或 BACF 中的裁剪矩阵，因此可以直接利用相关滤波基本算法中的闭合解的形式对模型求解，从而克服了迭代求解造成的时间消耗，算法效率得到明显提升。

(4) 连续卷积操作

C-COT (Continuous Convolution Operators for Visual Tracking) [37] 利用了多卷积层深度特征，并创新性地提出了连续空间域插值操作来应对不同卷积层特征分辨率不同的问题。如图 9 所示，模型通过频域上的隐式插值操作将特征图插值到连续域，用于集成不同分辨率特征，算法在连续域上表征和计算相关滤波器，获取了更高的目标定位精度。C-COT 也连续两年获得视频目标跟踪挑战赛 VOT-2016，VOT-2017 的第一名。尽管 C-COT 在跟踪精度上取得了优越的成绩，但是其算法速度只有在 GPU 上的不到 1 fps，与目标跟踪的实际应用差距较大。在此基础上，ECO (Efficient Convolution Operators for Tracking)[38] 通过卷积因式分解运算减少模型参数，生成样本空间模型减少样本数量，模型稀疏更新策略减少更新频率三个方法，有效地提升了算法的运行速度。之后，UPDT[41] 也是在 ECO 的基础上进一步挖掘深度特征的潜力（如 3.1.2 所述），使得算法效果进一步提升。

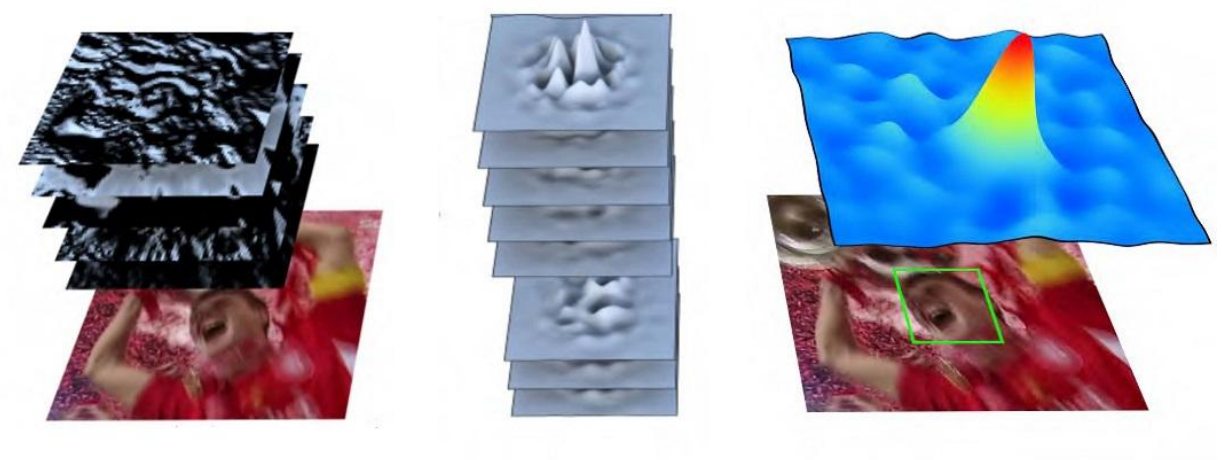

图 9. CCOT 算法特征和滤波器的连续化操作

通过对上述基于相关滤波的目标跟踪算法的分析，可以看到目前性能表现最好的算法，例如上述的 ASRCF，UPDT 算法，往往都使用了更具表达能力特征以及更加复杂的模型。综上，不难发现，开发更强有力的特征（从最早的单通道灰度特征，到多通道梯度特征，再到颜色特征，以及目前的深度特征），建立更加优化的模型（如改进滤波器更新方式，改进尺度估计方法，缓解边界效应），是基于相关滤波目标跟踪算法的近年来最主要发展趋势。

### 3.2 基于孪生网络的目标跟踪方法

#### 3.2.1 全卷积孪生网络

孪生网络 (Siamese Network)[51] 是指具有相同或相似结构的两个并行网络，基于孪生网络的相关算法早在 20 世纪 90 年代就被应用于模板匹配、相似度量等任务。由于孪生网络的参数较少，运行速度较快，被应用于很多其他任务。孪生网络最早用于视频目标跟踪任务是在发表于 2016 年的 SINT[52] 和 SiameseFC[24]，算法首次将目标跟踪问题转化为给定模板与候选图像的匹配问题。最早的

SiameseFC 算法就达到了较高的跟踪精度，并且维持了超高的算法运行速度（86 fps），也为后续的系列方法提供了良好的基础。图 10 展示了 SiameseFC 算法的网络结构，上层分支 $z$ 表示目标模板图像，由视频序列第一帧给定的目标区域生成，下层分支的输入是当前帧搜索区域，$x$ 表示搜索区域内部不同的目标候选图像，$z$ 和 $x$ 经过相同的特征映射操作 $\varphi$ 将原始图像映射到特征空间，得到具有相同通道数的特征向量，最后经过卷积操作 * 得到响应图，其中各个位置的值代表不同目标候选图像与目标模板图像的相似度，通过取最大值选择最相似目标候选区域，完成目标定位跟踪。上图中特征映射操作 $\varphi$ 由卷积神经网络实现，并且两个分支中 $\varphi$ 具有相同的网络结构，因此称为孪生网络，而且在 SiameseFC 算法中网络结构中只包含卷积层和池化层，因此其也是一种典型的全卷积孪生网络（Fully-Convolutional Siamese Network）。

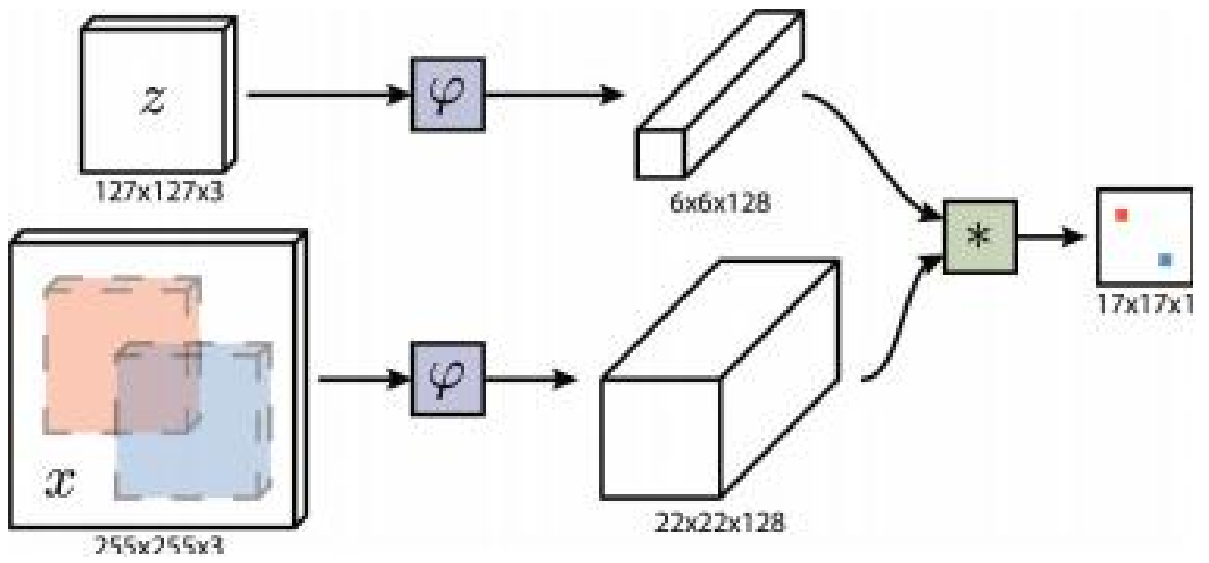

图 10. 全卷积孪生网络 (SiameseFC) 目标跟踪算法

### 3.2.2 区域候选孪生网络

SiameseFC 算法用于视频目标跟踪后，由于其网络结构简单，算法速度较高，引起相关学者的广泛关注和大量研究，其中典型的代表是 SiameseRPN[53] 系列算法。SiameseRPN 算法以 SiameseFC 为基础，将目标检测经典算法 Fast R-CNN[54] 的 RPN (Region Proposal Network) 即区域候选网络提取目标候选框的思想融入其中。如图 11 所示，相当于在待检测搜索区域上提取 $k$ 个目标候选框，网络上半部分与 Fast R-CNN 的分类网络类似，得到 $k$ 个目标候选框的响应图，再选取最高的响应值确定目标位置，下半部分与 Fast R-CNN 的回归网络类似，得到目标候选框与目标真实标定框的坐标差，作为补偿量修正检测结果。RPN 可以产生不同比例的候选框，因此很大程度上解决了目标跟踪问题中物体严重形变的问题。在 SiameseRPN 的基础上，DaSiamRPN[55] 通过训练集数据增广提高模型的泛化能力，还通过引入不同困难度的负样本训练提升模型的判别能力。最近的 SPM[56] 和 C-RPN (Cascade RPN)[57] 算法都是多阶段的 SiamRPN 扩展，其中 SPM 引入了经典的目标检测方法 Faster R-CNN[58] 的思路到目标跟踪网络，而 C-RPN 则借鉴了目标检测领域的级联网络 Cascade R-CNN[59] 的思想。SiameseRPN++[60] 和 SiamDW[61] 两个工作围绕如何在目标跟踪方法中使用更加深层的主干网络问题展开研究。SiameseRPN++[60] 改进了 SiameseRPN 中的样本采样策略，防止出现正样本都位于图像中心而影响目标定位的问题，在相关数据集上表现出良好的跟踪精度和鲁棒性。SiamDW[61] 研究了如何在孪生网络目标跟踪算法中利用更深和更宽的卷积神经网络提升算法的鲁棒性和精度。文章分析了直接利用更深的网络不能提升算法性能的原因：1）增加神经元的感受野会减少特征的区分度和目标的定位精度；2）卷积网络填补操作会对定位造成偏差。为解决上述问题，SiamDW 提出新的残差模块用于消除填补操作的负面影响，此外还进一步搭建了新的网络结构控制感受野的大小和网络步长。模块应用于 SiameseFC 和 SiameseRPN 算法得到了更好的跟踪结果和实时的运行速度。基于上述两项研究可以看到，在目标跟踪方法中应用更深的主干网络进行特征提取，可以进一步发挥深度学习方法在目标跟踪中的效力。

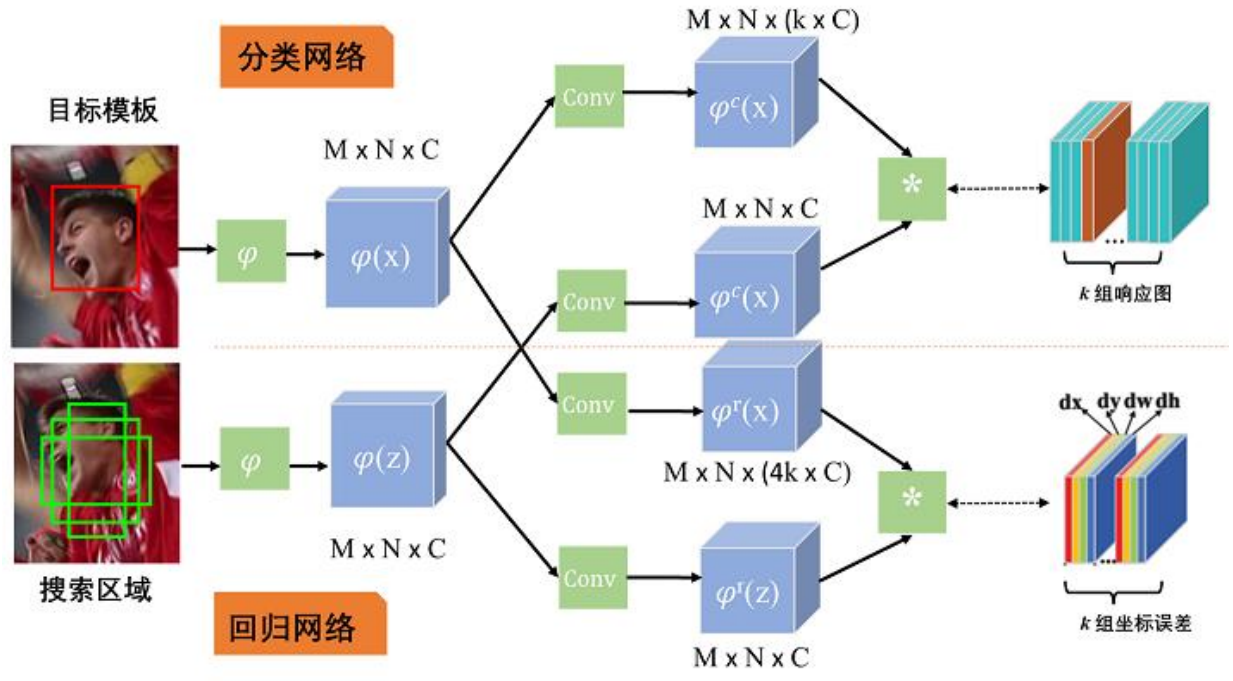

图 11. 区域候选孪生网络 (SiameseRPN) 算法

### 3.2.3 基于注意力机制的目标跟踪

RASNet[62] 将注意力机制模型引入孪生网络目标跟踪问题中，提出了通用注意力机制，残差注意力机制，通道注意力机制三种模型，其中通用注意力模型用于训练不同样本的共同特征，残差注意力机制用于捕获目标的形状外观等个体特征，而通道注意力机制用于建模特征通道的差异性，从而选

取更有效的特征。SiameAttn[63] 提出了可变形孪生注意力网络 SiamAttn (Deformable Siamese Attention Networks) 来提升孪生网络跟踪器的特征学习能力。SiameAttn 包括可变形的自注意力机制和互注意力机制两部分。自注意力机制通过空间注意力和通道注意力可学习到强大的上下文信息，并选择性地增强通道特征之间的相互依赖；而互注意力机制则可以有效地聚合与沟通模板和搜索区域之间丰富的信息。这种注意力机制为跟踪器提供了一种自适应模板特征隐式更新方法。文献[64]提出了像素引导的空间注意力模块和通道引导的通道注意力模块，以此来突出拐角区进行角点检测。注意力机制模块的引入提升了角点检测的准确性，从而提升了目标定位与矩形框检测的精度。

#### 3.2.4 基于时空关系的目标跟踪

DSiam[65] 引入了目标形变转换 (target variation transformation) 机制以及背景抑制转换 (background suppression transformation) 机制来应对目标在跟踪过程中的外观变化以及背景干扰等问题，提高了 SiameseFC 算法的鲁棒性。CLNet[66] 注意到 Siamese 网络测试时，面对一个新的序列，其目标和场景在离线训练中未出现，缺乏这些关键样本训练容易造成误判。于是作者提出了决定性样本 (decisive samples) 的概念，在跟踪过程中充分利用第一帧的目标标注信息来训练分类网络，另外，CLNet 不仅考虑了标注框内的模板信息而且考虑了其周围丰富的上下文信息用于调整，提升算法跟踪全新目标时的鲁棒性。文献[67]引入背景运动和轨迹预测模型，利用目标的历史轨迹对其当前和未来的空间位置进行预测，辅助目标跟踪过程中的时空定位。此外，还有一些算法，如 MemTrack[68, 69]，利用循环神经网络 RNN (Recurrent Neural Network) 建模目标的时空关系，本文将在后续章节 3.3.2 介绍。

#### 3.2.5 无锚点孪生网络目标跟踪

最近，关于基于 Siamese 网络的目标跟踪方法，一个重要的优化方向就是从锚点 (anchor) 方面改进网络。基于锚点的 (anchor-based) 网络，如 SiameseRPN，对锚点的数量、尺度、纵横比很敏感，同时训练网络还需要很多超参数，因此研究无锚点的孪生网络跟踪算法引起领域的关注。其中，SiamFC++[70] 增加了平行于分类分支的质量评估分支去除锚点的先验。类似的，SiameseCAR[71] 添加了一个与分类分支平行的中心突出分支对分类进行处理。SiamBAN[72] 将分类分支分为 2 个通道，预测响应图上每个点对应是前景还是背景，回归分支预测映射到原图上的每个点与标注框四条边之间的偏移量，这种回归分支的设计解决了基于锚点架构中分类与回归不一致的问题。Ocean[73]受启发于目标检测算法 FCOS[74]，提出了基于特征对齐的无锚点目标跟踪网络。由于无锚点网络无需锚点设计先验，同时可以更好地处理目标形状不规则与不固定的问题，因此也是今后的重要研究方向之一。

#### 3.2.6 其他孪生网络跟踪算法

此外，还有很多其他方法基于 SiameseFC 算法进行改进。例如，在网络结构方面，SA-Siam 建立了双路孪生网络用于视频目标跟踪，网络由语义分路与外观分路组成，每个分路均为用于相似性度量的孪生网络。SA-Siam 算法同时开发了目标语义信息与外观信息作为互补特征用于目标跟踪任务，在训练过程中同时保持了两类特征的异质性。在损失函数方面，SiamFC-tri[75] 则将三元组损失函数 (Triplet Loss) 引入孪生网络目标跟踪算法，利用目标样本，正样本和负样本组成的三元组用于训练，同时考虑了目标与正负样本的距离关系，提升了孪生网络目标跟踪算法的精度同时维持了算法速度。

### 3.3 其他深度学习目标跟踪方法

#### 3.3.1 卷积网络跟踪算法

MDNet (Multi-Domain Convolutional Neural Networks)[22] 是早期的完全基于深度卷积经网络 CNN 的目标跟踪算法，也是 VOT-2015 挑战赛的冠军算法。MDNet 通过在大量视频跟踪标注的数据集上训练得到了可用于通用目标表示的卷积神经网络，如图 12 所示，MDNet 包含共享层 (shared layers) 和多通道特定域层 (domain-specific layers)，这里域对应了训练集中不同的视频序列，而每一个通道负责在该域上通过二分类任务确定目标位置。另外，在共享层，MDNet 通过在每个域上的迭代训练来获取更一般的目标表示。在测试序列上，MDNet 组合共享层和预训练的二分类卷积层并在线更新构建新的网络，通过目标上一帧位置周围随机选取区域候选框，利用二分类网络判定该区域是否为目标，实现在线跟踪任务。MDNet 虽然取得了比较好的跟踪精度，然而仅有在 GPU 上约 1 fps 的运行速度，算法无法实际应用。在此基础上，RT-MDNet (real-time MDNet)[76] 使用了特征提取过程加速和实例分类模型优化的技巧，还通过全新的

损失函数学习得到了更好的目标建模方式，以区分目标与其具有相近语义的其他物体。通过将上述技巧与 MDNet 算法整合，新的算法达到了 25 倍的速度提升和与 MDNet 相接近的跟踪精度。

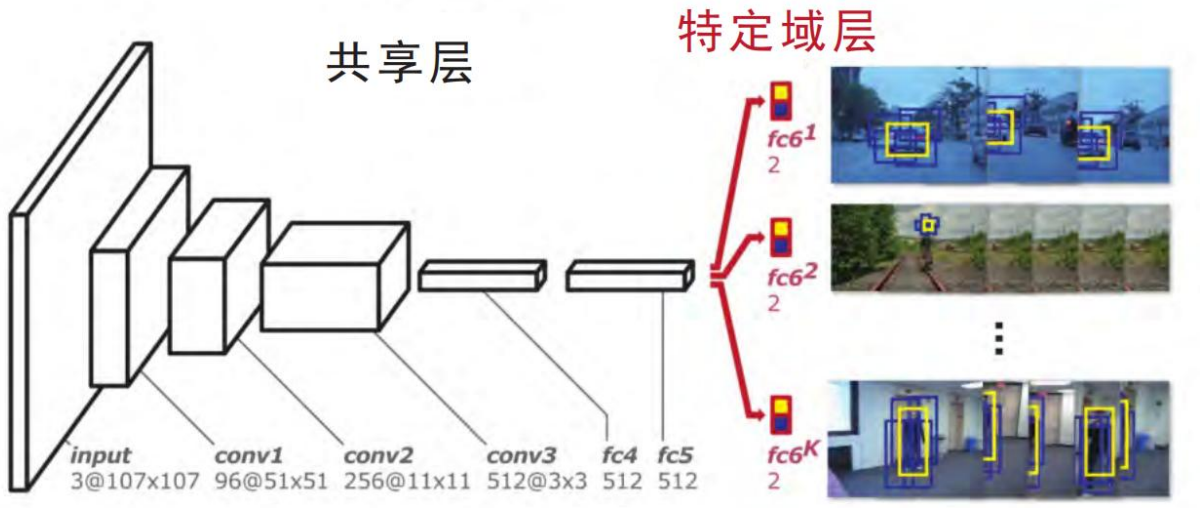

图 12. MDNet 算法示意图

TCNN (CNNs in a Tree Structure) [77]利用多个 CNN 模型，并构建成树结构用于目标跟踪任务，是 VOT 2016 年的冠军†。TCNN 算法最重要的出发点是模型的可靠性问题，即当目标被遮挡或丢失的情况下，利用当前状态更新的模型其实已经发生漂移 (model drift)，使用这样的模型也无法完成后续的跟踪任务。TCNN 利用树结构维持多个 CNN 并对每一个 CNN 模型进行了可靠性评估，还通过平滑的更新来保存模型的可靠性。因此，TCNN 算法在模型鲁棒性方面取得了明显的提升，达到了良好的跟踪精度。然而由于需要维持多个网络，因此算法的速度只有约 1 fps。

CREST (Convolutional Residual Learning for Visual Tracking) [78] 首次将残差学习的概念应用到目标跟踪算法。之前的深度目标跟踪网络通过回归获得二维高斯响应图，计算峰值确定物体的位置。而当目标发生形变，背景干扰等情况时，网络将无法预测的准确二维高斯响应，使得峰值发生偏移，并累计误差，导致物体位置跟踪失败。CREST 将单层卷积作为基础映射 (base mapping) ，引入了残差映射 (residual mapping) 机制来捕获基础映射和真实高斯响应之间的差异。在基础映射的输出和真实高斯响应接近时，残差网络几乎不发生响应。在物体发生变化时，残差网络发生响应捕获基础映射输出与真实高斯响应之间的差异，并通过求和操作补充基础映射输出，使得整个网络的输出更接近真实结果，更精确地定位目标。

综上不难看出，上述基于卷积神经网络的目标跟踪算法在该任务上并未发挥出明显的优势。上述方法最大的不足在于算法速度过慢，无法满足目标跟踪任务的实时性需求。这主要是因为深层卷积神经网络往往结构复杂，计算较慢。同时，深度的卷积网络可以更好地实现目标的语义级表达，但是这类优势在目标跟踪任务中并无法充分展现。因此，未来卷积神经网络用于目标跟踪任务可以向轻量级网络发展，并引入目标检测中的单样本学习 (one-shot learning) 思想，在极少训练样本的基础上实现目标的可靠而鲁棒的建模。

3.3.2 循环神经网络跟踪算法

为了更好地捕获目标跟踪过程中目标的外观变化从而提升跟踪精度，有研究者考虑引入循环神经网络 RNN 模型。RFL[79] 算法直接将视频中的目标序列送入 RNN 网络用来训练适应于特定目标外观变化的跟踪滤波器。与之前基于目标检测的跟踪算法需要在线调整网络参数以适应目标变化的方式不同，当 RFL 网络离线训练完成后，在线跟踪测试过程中不需要针对指定目标进行网络参数的微调，这使得算法具备更强的实用性。MemTrack[68] 提出了基于长短期记忆模型 LSTM 的动态记忆网络以适应目标跟踪过程中外观变化的情形。与 RFL 类似，MemTrack 在测试过程中可以实现完全利用前向传播计算结果并利用更新额外的记忆网络适应目标变化，而不需要在线调整网络参数。另外，之前的网络通过离线训练后模型能力保持不变，不同的是，MemTrack 的模型能力会随任务的记忆需求的增加而加强，因此可以更好地用于长时间目标跟踪问题。

循环神经网络在目标跟踪任务中有一定的应用，其主要优势在于长时间目标跟踪任务。然而，目前 RNN 用于目标跟踪任务还远远不足。例如，对被跟踪目标的长期记忆未能考虑其时序变化，因此对变化频繁的目标跟踪不利。另一方面，对目标的短期记忆容易受目标遮挡等情形的影响而产生模型漂移。为此，利用 RNN 模型更好地综合目标的长期稳定性特征（如语义特征）和短期可变性特征（如颜色尺寸），将能进一步发掘其在目标跟踪任务中的潜力。

3.3.3 图卷积网络跟踪算法

GCT (Graph Convolutional Tracking) [80] 首次将图卷积网络 GCN (Graph Convolutional Network) 应用到目标跟踪问题。考虑目标时空外观特征和环境特征对目标跟踪有很大帮助，但现有的目标跟踪

† VOT-2016 挑战赛的第一名是 C-COT 算法，但由于 VOT 挑战赛规定冠军算法作者不能包含组委会成员，因此当年比赛冠军是获得第二名的 TCNN 算法。

算法大多没有充分利用不同环境下目标时空域上的外观建模。因此，GCT 算法在 Siamese 算法框架下，综合考虑了两类图卷积网络模型用于目标外观建模，包括时空域 GCN (spatial-temporal GCN) 构建目标时序性的结构化的特征，以及环境 GCN (context GCN) 利用当前帧目标上下文信息构建目标自适应特征，提高了模型对目标前背景的建模能力，提高算法的鲁棒性和精度。

目前，图卷积神经网络在目标跟踪领域的应用还很少，不过图卷积网络具备的强大关系表达能力对建模目标与周围环境之间的关系具有特殊的优势。未来，如果能进一步利用图网络表达目标与场景其他物体（特别是外观相似的同类物体）的空间位置及分布关系，以及目标的在视频序列中的时空位置变化，将有望为目标跟踪带来新的可能性。

3.3.4 重叠预测网络跟踪算法

ATOM[81] 引入一种新颖的多任务跟踪框架，主要包括两个成分，用于目标估计和分类。作者受启发于 IoU-Net[82]，训练一个目标预测模块预测估计的目标框与真实结果之间的重叠率（交并比），目标是最大化每一帧预测的重叠率，目标预测模块在大型数据集上进行离线训练。此外，ATOM 还设计了一个分类模块，并且是在线训练的，保证了较高的鲁棒性。最近，基于 ATOM 网络框架，DiMP[83] 提出了判别力强的损失函数指导网络学习更加鲁棒的特征以及更有力的优化器加快网络收敛。在此基础上，PrDiMP[84] 进一步利用概率回归模型，融合了基于置信度回归方法的优点，使得目标位置回归更加精确。

ATOM 系列重叠预测网络跟踪算法兼顾了 Siamese 网络为代表的大规模数据离线训练方法，以及自适应前背景变化的在线更新模式，在精度和速度上都具有很好的效果，也有望成为深度目标跟踪算法的重要框架。该系列算法也启发未来深度跟踪算法的检测器应进一步考虑在跟踪过程中对目标变化、背景移动、目标遮挡等信息进行时序建模与在线更新。

### 3.4 解决目标跟踪中的挑战问题

3.4.1 目标消失问题

目标消失（包括遮挡，出视野）是目标跟踪问题最困难的挑战之一，也有很多算法尝试解决目标丢失对跟踪造成的影响。LMCF[85] 从如何判断跟踪结果是否准确这个问题入手，提出了高置信度模型更新策略。传统的相关滤波 (CF) 算法不判断当前跟踪结果是否可靠，在每一帧都对 CF 进行更新。但是，在目标被遮挡时对 CF 实施更新操作，会造成跟踪器产生模型漂移 (model drift)，目标重新出现后也无法识别。通过实验验证作者发现跟踪响应图在被遮挡时变化剧烈，可以通过跟踪结果响应图的可靠程度和变化趋势来避免错误更新。对此，作者提出了一个新的指标— APCE (average peak-to-correlation energy) 用于反映响应图的变化程度。文章指出，当 APCE 值突然减小，造成的原因可能是目标丢失或被遮挡的情形发生，此时则不对模型进行更新，从而一定程度上避免模型漂移产生。LMCF 算法只有在当前 APCE 值大于历史均值条件满足才对模型进行更新，算法不仅缓解了模型漂移问题，还降低了模型更新频率，达到了算法加速的效果。类似地，SSRCF (Selective Spatial Regularization for CF tracking)[86, 87] 采取了三种模型更新策略，在目标发生轻微变化时采用一般模型更新，在目标几乎不发生变化时不更新模型，在目标发生严重遮挡时采用背景更新模型，即通过学习目标背景特征辅助定位目标。算法利用了马尔科夫决策过程实现上述三种模型更新方式的切换，有效地提升了目标遮挡情形下算法的鲁棒性。然而，面对复杂的目标消失问题，仅靠模型更新似乎应对乏力。最近，Siamese R-CNN[88] 利用目标重检测机制解决跟踪中目标消失问题。Siamese R-CNN 首先提取第一帧中的目标区域作为参考，在当前帧利用候选网络生成目标的候选框，在搜索图像中寻找与目标最相似的物体，并与上一帧得到的候选框进行匹配，根据距离判断时候是同一目标，再将相邻两帧图像块距离较近的目标一起送入重检测网络中再次判断是否是同一目标。将检测到的目标通过关联的方式形成跟踪轨迹链，如果一但有干扰目标存在，那么开辟一条新轨迹，最终得到感兴趣目标的最终轨迹。因此，Siam R-CNN 能够有效应对长时间目标跟踪，能够在目标消失后重新出现时重新检测到跟踪目标。另外，文献[67]提出 BM Net (background motion network) 利用目标历史轨迹解决目标遮挡过程中外观信息不可靠的问题。当目标遭受严重遮挡时，当前帧的空间域中几乎没有有用的信息来检测目标物体。因此，当跟踪器由于遮挡而丢失目标时，根据目标的过去轨迹来预测的后续目标的位置比依靠低基于模板匹配的跟踪器预测更加可靠。为提高轨迹预测能力，BM Net 提出了三个先验：1）摄像机运动对物体运动轨迹的影

响很大；2）物体的过去轨迹可以作为显著线索来估计物体在空间域中的运动；3）先前的帧包含目标对象的周围环境和外观，这有利于预测目标的位置。基于上述先验和想法，BM Net 采用了多流卷积 LSTM 网络用于长期被遮挡的目标跟踪，有效地缓解了目标长时间遮挡对跟踪的影响。如图 13 所示，其中绿色框表示跟踪目标，红色框表示 BM Net 算法的跟踪结果。可以看到，BM Net 通过相机运动，目标运动和背景运动建模，对目标的轨迹进行预测（如图红色曲线），有效地应对目标跟踪过程中发生的长时间遮挡（如图第一行运动员头部）和严重遮挡消失（如图第二行小女孩）。目前，以 Siamese 网络为基础的大量目标跟踪算法，由于其对目标初始化（第一帧）外观信息的强大建模能力，针对目标消失问题已经具备一定的鲁棒性。然而，当目标消失又重新出现后，其外观特征与目标初始帧或目标消失前发生较大差异的情况，将会为目前的目标跟踪算法带来较大的挑战，也是目前目标消失问题需要进一步解决的关键地方。

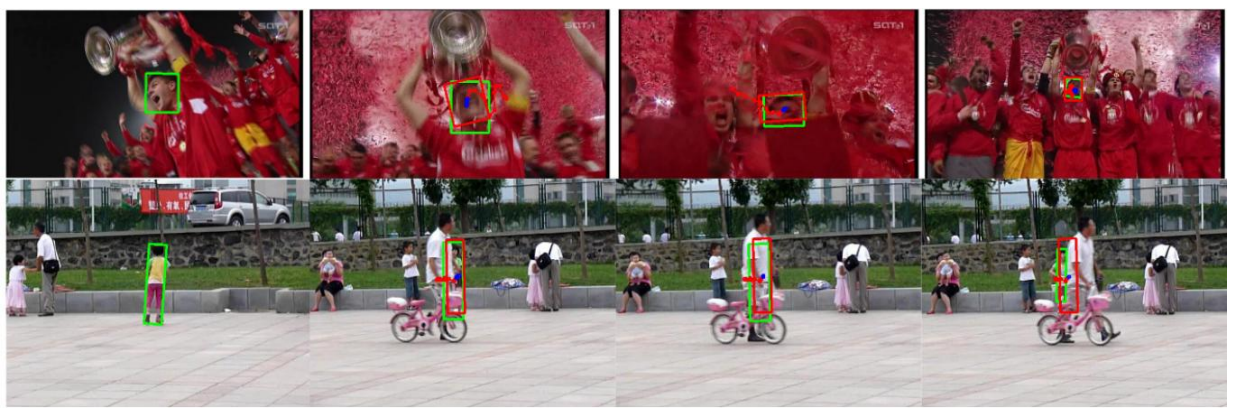

图 13. BM Net 算法解决目标遮挡消失问题

### 3.4.2 目标形变问题

为解决目标跟踪问题中目标形变（非刚体形变，尺度变化）的问题，CSR-DCF[33] 利用了基于统计概率模型的空间权重图使相关滤波系数更好地适应不同形状的前景目标。DSAR-CF[25] 则考虑了显著性检测方法获得目标精确的形状轮廓信息，用于辅助提升相关滤波算法中的空间正则化方法的效果。两类算法都在相关滤波目标跟踪算法的框架下有效地缓解了目标形变对跟踪造成了挑战。在孪生滤波 Siamese 算法框架下，RASNet[62] 利用离线训练的目标自适应注意力网络获得物体的外观形状信息，从而辅助深度网络提取目标特征的表示，提升了表征学习和鉴别学习能力，进而提升了网络的前背景区分性和自适应性。SiamMask[89]、D3S[90] 等方法将目标分割引入目标跟踪，精确获取目标前景区域，以便更好应对跟踪问题中目标的复杂形变。例如，D3S 基于目标跟踪网络得到比较精确的目标位置信息，然后结合目标分割的方法得到目标的分割前景、背景特征响应图，最后将目标位置信息与分割的响应图进行融合，进行上采样，同时得到目标跟踪与分割的结果。网络简单有效，分割结果精度高，在线跟踪的鲁棒性强。以 D3S 算法为例，图 14 展示了算法对非刚性易变形目标的跟踪结果。可以看到，即使目标，如身体、手掌等发生严重的形变，D3S 算法不仅可以准确地跟踪目标，还能实现目标的像素级分割获取其轮廓。尽管目标形变问题已经靠结合目标分割任务得到一定程度的解决，其仍是目标跟踪问题的主要挑战。目前，该问题最难的地方在于目标发生的快速严重形变，即目标外观短时间内变化剧烈，从而使得目标跟踪模型无法做出及时调整。同时，目标形变往往与其他挑战，如目标快速移动、目标遮挡等同时发生，更加重了其困难程度，也是今后目标跟踪任务需要着重解决的问题之一。

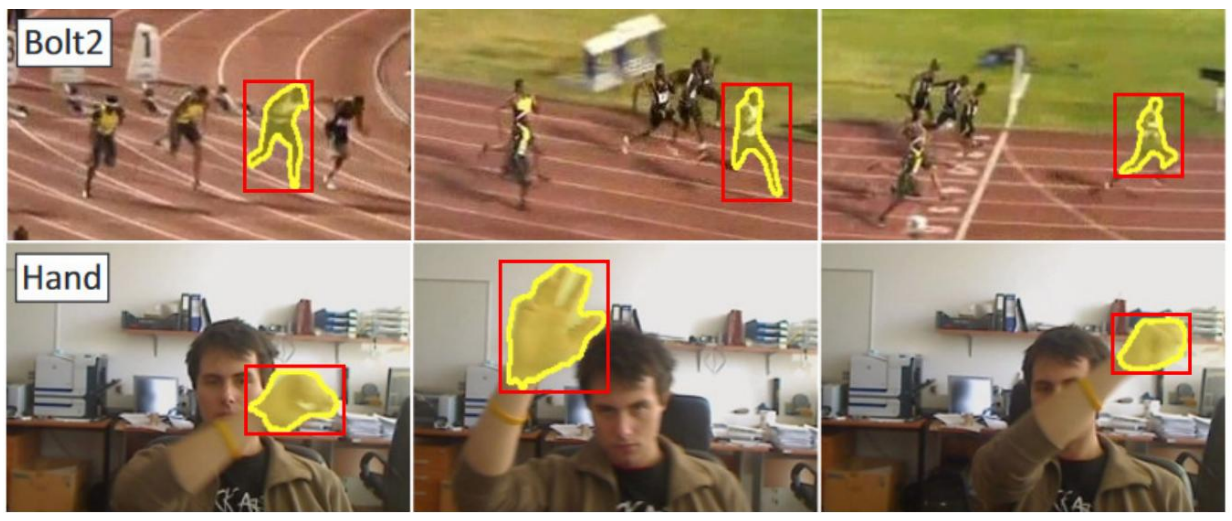

图 14. D3S 算法解决目标形变问题

### 3.4.3 背景干扰问题

面对目标跟踪中背景干扰的问题，许多方法提出利用目标周围上下文 (context) 信息来缓解目标周围相似物体或复杂背景对目标跟踪造成的影响。例如，CACF (Context-Aware CF)[91] 利用目标周围区域采样，显式地学习了目标周围的背景信息，并将这种框架广泛地用到多种基于 CF 的跟踪算法上，得到了效果的显著提升。为解决相似物体干扰的问题，LMCF[85] 则提出了多峰二次重检测机制。在目标附近有特征相似的干扰目标出现时，目标跟踪算法会产生包含多个峰值（极大值）的响应图，而峰值最大的位置有可能并非目标而是干扰物体，从而使跟踪结果发生错误。多峰目标检测是指当多个峰值与最高峰值的比例大于给定阈值时，对多峰值进行二次目标检测，最后所有响应图的最高点为最终目标定位结果。Siamese 网络同样面临背景或相似目标干扰的问题，PGNet[92] 对此进行了详细的讨论。Siamese 网络中模板特征与检测特征匹配区域的坐标直接映射到搜索图像中，可以得到与目标具有相同尺度的理想匹配区域。然而，由于卷积网

络感受野的影响，随着网络深度的增加，特征区域可能对应着输入图像中远大于目标的区域。这时就引入了大量的背景信息，甚至会覆盖目标的一些特征，使目标与背景中相似的地方难以区分。为此，PGNet 定位子网络由 3 个 PGM (Pixel to Global Matching) 模块组成，PGM 用像素到全局匹配来进行相似度计算，每个 PGM 产生一组基于锚点的分类和回归结果。PGNet 分别提取浅层、中层和深层特征作为目标定位子网中相应的 PGM 的输入。PGNet 减少了背景的干扰，并在降低计算成本的同时提高模型精度和鲁棒性。最近，Bhat 等人在 KYS (Know Your Surroundings) 算法[93]提到目标跟踪不仅需要对跟踪目标本身进行外观建模，还需要考虑到目标周围场景物体的存在和位置，场景信息可以用于排除错误的目标候选区域以避免相似物体干扰。如果不考虑目标环境信息而仅考虑目标本身信息，在相似目标出现时跟踪器很难鉴别待跟踪目标，如图 15。因此，文献[93]提出了利用目标场景信息的目标跟踪算法架构，该方法将场景信息建模生成稠密的局部状态向量，用于将局部区域编码为目标、背景以及干扰物体，从而将该状态向量与目标外观模型整合用于跟踪定位目标。图 16 展示了最新的针对目标相似物体干扰的 KYS 算法在目标背景干扰场景下的跟踪结果，其中蓝色表示基准方法，红色表示采用 KYS 中的背景建模后的方法。可以看到，周围存在相似物体对特定目标跟踪具有严重的干扰，基准方法无法实现目标与干扰物的区分。如图，在考虑目标与背景物体的综合信息后，KYS 算法很好地实现了对同类外观近似目标的有效区分与持续跟踪。尽管如此，目前的算法针对大量密集相似目标同时出现的场景仍无法全部解决，例如 VOT 挑战赛中的视频序列在蚂蚁群中跟踪某只蚂蚁，在落叶堆中跟踪某片落叶的场景，仍然极具挑战。

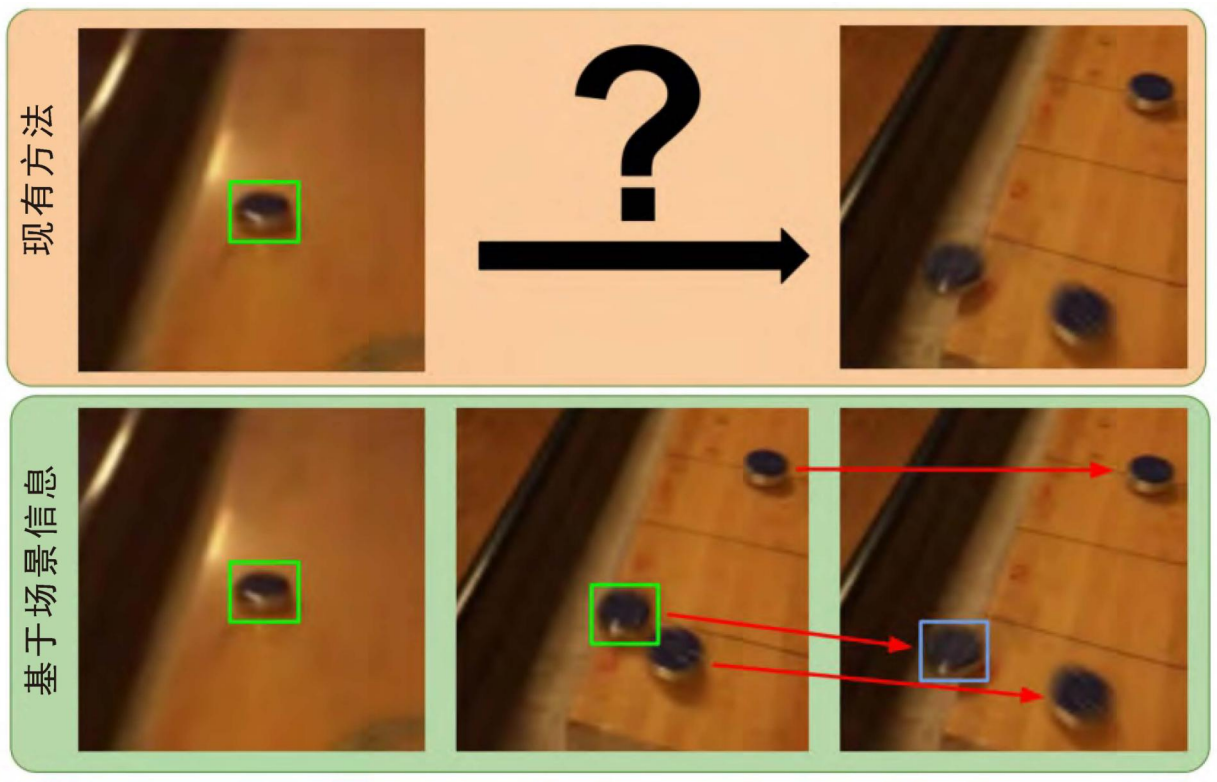

图 15. KYS 算法利用目标环境信息进行跟踪

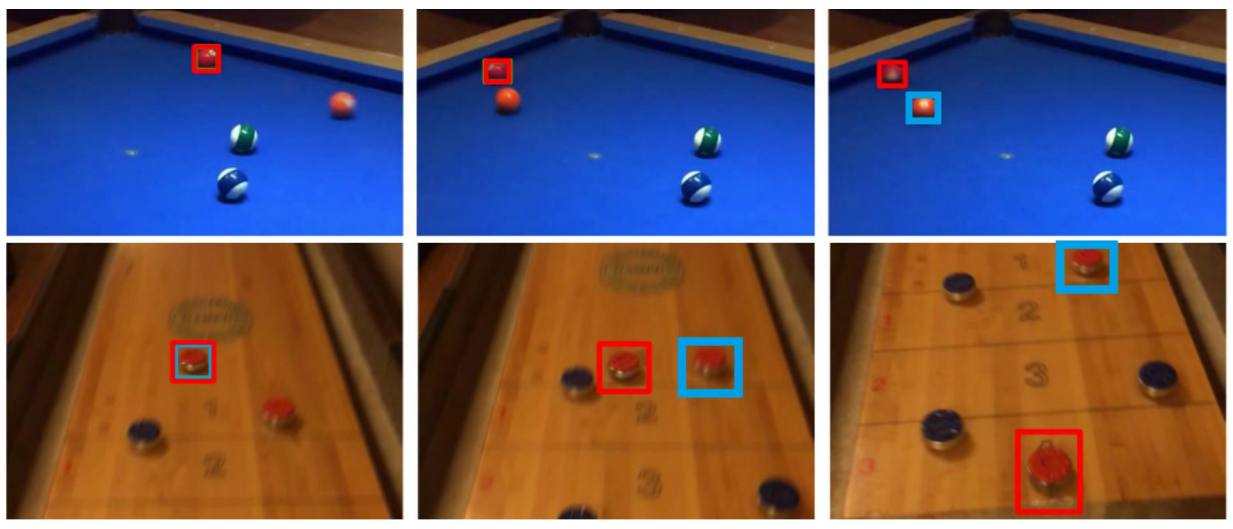

图 16. KYS 算法解决目标相似问题

### 3.4.4 目标移动问题

目标移动也是目标跟踪面临的挑战之一，其对目标跟踪带来的主要影响是目标快速移动可能导致其相邻两帧的位置差异较大，从而使得目标的待搜索区域无法确定。区域过小则可能无法包含目标，区域过大则导致引入较多的干扰信息。MACF (Motion-Aware CF)[94] 提出包含目标运动的搜索区域选择策略，该方法通过估计的 3D 运动信息，用于预测目标可能出现的位置，解决上述问题。目标移动带来的另一个影响就是运动模糊，其同样可能对目标跟踪带来影响。早期的工作[95]试图对运动模糊的进行内核估计从而减低模糊对跟踪的影响。后续工作[96, 97]通过稀疏表示的方法构建对运动模糊鲁棒的跟踪器。近来，文献[98]则提出对运动模糊建模来实现对目标跟踪效果的提升。目标的移动虽然为目标跟踪带来了挑战，同时也可以带来新的信息。目前主流的目标跟踪算法大多依赖于目标的外观信息，然而，目标的移动同时也可以引入新的特征，即目标运动特征。DMF[99] 首次提出将目标的深度运动特征利用于目标跟踪任务，并发现目标的运动特征可以为目标跟踪提供有区分度且与外观特征互补的信息，进一步提升运动目标跟踪的效果。最新的文献[100]通过大量试验论述了运动模糊对视频目标跟踪的影响，并得出有趣的结论，轻微的运动模糊不仅不会对跟踪造成负面影响，反而对目标跟踪带来帮助。文献[100]进一步探讨了目标快速移动产生运动模糊时去模糊算法对目标跟踪算法的影响。如图 17 所示，上方序列展示了经典目标跟踪算法 ECO[38] 的跟踪结果，可以看到在目标快速运动产生运动模糊时，跟踪器丢失目标。下方序列展示了在原图上利用图像去模糊算法处理后视频上的跟踪结果，可以看到跟踪器可以持续跟踪目标。目前，针对目标运动，特别是目标运动模糊对跟踪算法的影响研究相较于其他类型的挑战还比较少。尽管上述实验表明去模糊算法的预处理可以一定程度上缓解目标运动模糊对跟踪带来的挑战，

但是事先处理的方法一定程度上限制了其实用性。另外，文献[100] 也表明并非去模糊算法的使用一定对目标跟踪算法有利。因此，目标快速运动对目标跟踪的影响，以及如何高效统一地处理运动模糊对目标跟踪带来的挑战，还有待进一步探索。

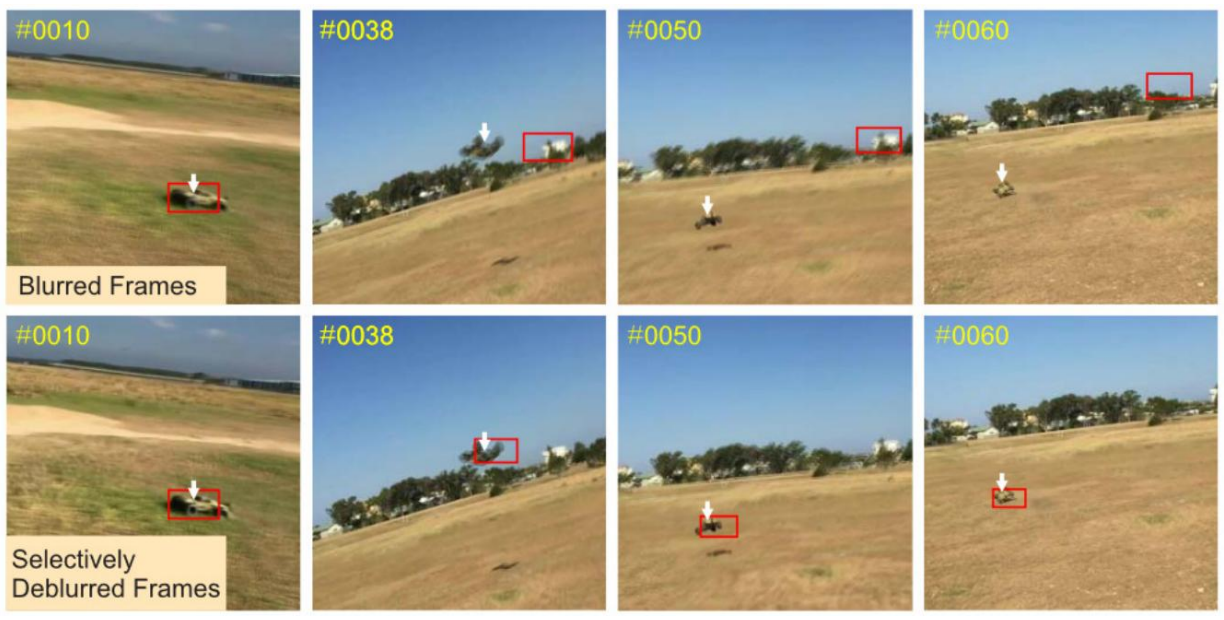

图 17. 去模糊算法与目标跟踪

### 3.5 目标跟踪方法小结

回顾近年来视频目标跟踪的发展，可以分析总结得到以下几点：

1) 相关滤波 CF 目标跟踪方法的优点：i) 在线更新，可以更好地适应于目标的时序变化；ii) 无需训练，不需要大量的标注数据离线训练。

相关滤波 CF 目标跟踪方法的不足：i) 模型在更新过程中容易漂移，一旦目标丢失将导致滤波器被污染；ii) 无法高效率用基于大规模数据训练的深度特征提高算法精度。

2) 相关滤波 CF 目标跟踪方法的改进方案：i) 进一步发挥深度特征的优势，将更有效的特征与 CF 算法框架结合；ii) 与 Siamese 网络等方案的思想相结合，在保持目标在线更新的过程中更有效地保留目标初始（第一帧）信息。

3) Siamese 网络目标跟踪方法的优点：Siamese 系列算法可以保留目标第一帧模板数据，大量的离线训练可以使其充分地发挥深度学习的优势。

Siamese 网络目标跟踪方法的不足：Siamese 网络无法真正意义上抑制背景中的困难样本，也就是说离线的学习从本质上无法区分长相相似的目标，如行人、车辆的实例级区分。不能像 CF 相关算法通过分析整个环境的上下文关系来进行调整，其网络模式更像是单样本目标检测任务。

4) Siamese 网络等深度学习跟踪算法的改进方案：i) 集成在线学习的 Siamese 网络，可以为提升目标跟踪鲁棒性提供新的方向；ii) 当前的目标跟踪特征网络大多基于主流主干网络，如 ResNet 等，而开发特定结构的特征提取网络用于目标跟踪任务，如借鉴单样本学习 (one-shot learning) 中的特征提取网络，可以提取更适合的特征；iii) 当前的大规模数据集，如 LaSOT 等，其训练数据与测试数据的目标类型高度重复，导致模型更偏向于跟踪固定类别的目标。而如何提升模型的泛化性能，是研究通用非特定目标跟踪的重要基础；iv) 现有的 Siamese 网络算法结构日益复杂，如何实现网络结构小型化是算法落地的重要需求。

5) 目标跟踪方法框架发展趋势：从最近出现的跟踪算法来看，算法发展方向呈现多元化发展趋势，更多新的网络结构，如图网络、循环神经网络、动态记忆网络，和方法技巧，如注意力机制、无锚点策略、上下文关系等，开始应用于目标跟踪任务。其他深度目标跟踪方法虽然呈多样化趋势发展，但是并未形成完整体系。近年来的主流框架仍然为基于相关滤波和孪生网络的方法。另外，基于 ATOM[81] 及其系列算法 DiMP[83], PrDiMP[84] 综合了深度学习离线训练与 CF 在线更新的优势，在算法精度上表现突出。从未来近几年内目标跟踪方法的发展趋势来看，CF 目标跟踪方法已经发展地比较成熟，未来拓展空间有限。基于卷积神经网络的深度学习算法，特别是 Siamese 框架下的目标跟踪算法仍是主流。此外，无论是早期的 CF 算法还是后来的的 Siamese 网络，都无法有效应对长时间目标跟踪任务，因此面向长时间目标跟踪的深度学习算法框架将对目标跟踪领域具有重要意义。

## 4 视频目标跟踪数据集及结果分析

### 4.1 数据集统计分析

表 1 总结了目前视频目标跟踪主流的 13 个数据集的统计信息。本文主要介绍和比较相关数据集的规模（包括所含视频数目，所含帧数目）、时长（包括视频平均时长，数据集视频总时长）、多样性（包括数据集所含目标类别，数据集视频分类标签数）、及数据集其他属性（包括帧率，典型特点，和论文出处）。

OTB 数据集：OTB-2013[16], OTB-2015[20] 分别包含 50 和 100 个视频序列。其中 OTB-2013 是视频目标跟踪领域首个用于算法效果评估的基准数据集，OTB-2015 则是在 OTB-2013 的基础上进一步扩大数据规模，是视频目标跟踪领域使用最广泛，最经典的基准数据集之一。数据集共包含 10

表 1: 视频目标跟踪数据集属性统计

| 数据集 | 规模 | | 时长 | | 多样性 | | | 属性 | |
|---|---|---|---|---|---|---|---|---|---|
| | # 视频 | # 帧数 | 均时长 | 总时长 | 种类 | 标签 | 帧率 | 特点 | 出处 |
| OTB-2013[16] | 51 | 29K | 578 | 16.4mins | 10 | 11 | 30fps | 首次,经典 | CVPR13 |
| OTB-2015[20] | 100 | 59K | 590 | 32.8mins | 16 | 11 | 30fps | 经典 | TPAMI15 |
| TC-128[101] | 128 | 55K | 429 | 30.7mins | 27 | 11 | 30fps | 彩色 | TIP15 |
| UAV123[102] | 20 | 113K | 915 | 62.5mins | 9 | 12 | 30fps | 无人机 | ECCV16 |
| UAV20L[102] | 123 | 59K | 2,934 | 32.6mins | 5 | 12 | 30fps | 航拍,长时 | ECCV16 |
| NUS-PRO[103] | 365 | 135K | 371 | 75.2mins | 8 | - | 30fps | 行人,刚体 | TPAMI17 |
| NfS [104] | 100 | 383K | 3,830 | 26.6mins | 17 | 9 | 240fps | 高帧率 | ICCV17 |
| OxUvA [105] | 366 | 1,550K | 4,599 | 14.4hours | 22 | 6 | 30fps | 户外,长时 | ECCV18 |
| TrackingNet[106] | 30,643 | 14,431K | 498 | 141.3hours | 27 | 15 | 30fps | 户外,大规模 | ECCV18 |
| LaSOT[17] | 1,400 | 3,520K | 2,506 | 32.5hours | 70 | 14 | 30fps | 高质,大规模 | CVPR19 |
| GOT–10k[107] | 10K | 1.5M | 15 | 41.7hours | 563 | 6 | 10fps | 单次协议 | TPAMI21 |
| CDTB[108] | 80 | 101,956 | 1,274 | 56.5mins | - | 13 | 30fps | RGB-D | ICCV19 |
| RGBT234[109] | 234 | 234K | 1,000 | 130mins | - | 11 | 30fps | RGB-T | PR19 |

种类别的目标，整个数据集根据目标跟踪面临的常见问题共包含 11 类标签，包括：目标遮挡，视野消失，目标变形，尺度变化，面内旋转，面外旋转，背景扰动，光照变化，快速移动，移动模糊，低分辨率。每个视频根据其特点被标记为多个上述标签，OTB 对视频的属性划分也为后续的其他数据集提供了基础和参照。不过，随着视频目标跟踪算法今年来的快速发展，OTB 数据集用作效果评估已逐渐呈现饱和态势，因此更多大规模数据集应运而生。

TC-128 数据集[101]：与 OTB 数据集同时包含灰度视频和彩色视频不同，TC-128 数据集 128 个视频序列全部为彩色图像，专门用于评估基于颜色特征加强的目标跟踪算法，也是视频目标跟踪的常用数据集之一。TC-128 数据集包含 16 类目标，视频属性划分完全借鉴了 OTB 数据集的方式。

UAV 数据集[102]：UAV123 和 UAV20L 分别包含 123 个短视频和 20 个长视频序列，特定用于无人机目标的跟踪问题。

NUS-PRO 数据集[103]：共包含 365 个视频序列，主要用于行人和刚性物体目标的跟踪。NUS-PRO 是当时规模较大的基准数据集，该数据集的特点包括视频大多采用移动相机拍摄，并且在数据标注上不仅标注了目标的位置，还标注了遮挡程度。

NfS 数据集[104]：全称为 Need for Speed，旨在评估具有更高运行速度的跟踪算法。数据集提供了 1,000 个帧率为 240 fps 的高帧率视频序列，可用于精确分析目标外观变化对跟踪算法的影响。

OxUvA 数据集[105]：是牛津大学(Oxford University) 提出的由户外场景视频组成的大规模数据集，包含 366 个视频序列，总时长超过 14 小时。数据集构建的初衷是现有数据集视频时长较短且目标始终可见，与实际应用不符。所含视频平均长度超过 2 分钟并且频繁出现目标消失的情形。

TrackingNet 数据集[106]：是目前规模最大，所含帧数最多，时长最长的视频目标跟踪数据集。数据集包含超过 3 万的视频序列，超过 1,400 万个目标框标注。该数据集面向基于深度学习的目标跟踪算法，划分为训练集和测试集两部分分别用于算法的训练和基准测试，数据集还配套专门的在线评估服务用于相关算法的公平测试和比较。

LaSOT 数据集[17]：同样是最近提出的大规模长时间视频目标跟踪数据集，包含 1,400 个视频序列，涉及 70 类目标，数据集同样划分为训练集和测试集两部分。值得一提的是，LaSOT 数据集提供

了高质量和丰富的数据标注，目标框的标注-检查-微调机制使得标注更加精准，数据集还同时提供了视觉和语义的配套标注，方便该数据集拓展至其他任务的训练和评估。

GOT-10k 数据集[107]：基于著名的 WordNet[110] 结构建立而成，其包含了 560 类常见目标和 87 类运动模式。GOT-10k 数据集提供了 10,000 个视频片段和大约 150 万个目标标注框。值得一提的是，GOT-10k 数据集首次引入了单样本协议的训练测试方法 (one-shot protocol)，即训练集与测试集目标类别无重叠，其目的是更好地适配于非特定目标跟踪的任务。

CDTB 数据集[108]：是目前规模最大，最多样的 RGB-D 视频目标跟踪数据集，可用于评估基于深度-颜色信息的目标跟踪算法。数据集包含室内和户外场景下的视频序列，同时包含目标位姿变化，目标遮挡，视野消失等多种真实场景。

RGBT234[109]：是目前规模最大的 RGB-T 视频目标跟踪数据集，用于评估基于颜色信息加红外信息视频的目标跟踪算法。RGBT234 数据集中彩色视频与红外视频具备高精度的时间同步，不需要额外的前处理和后处理操作进行视频对齐。另外，数据及提供了遮挡程度的标注，可用于目标遮挡对不同跟踪算法跟踪精度的敏感性分析。

VOT 挑战赛[111-119]：除上述基准数据集外，VOT 挑战赛是视频目标跟踪研究领域最大规模，最具影响力的比赛。VOT 挑战赛自 2013 年开始，2014 年起固定每年在计算机视觉领域的顶级学术会议上 ECCV/ICCV 以研讨会的形式举办。近年来，VOT 挑战赛开始分为多个子赛事，侧重于不同特点的视频目标跟踪问题，包括 (i) VOT-ST：基于短时视频的目标跟踪；(ii) VOT-RT：具有实时性能的视频目标跟踪算法挑战赛；(iii) VOT-LT：基于长时视频目标跟踪；(iv) VOT-RGBD：基于 RGB 和深度图像视频的长时目标跟踪；(v) VOT-RGBT：基于 RGB 和热成像视频的短时目标跟踪。表 2 汇总了历年 VOT 挑战赛的赛事情况，其中短时视频目标跟踪是 VOT 挑战赛的经典项目，也是参赛最多，竞争最激烈的赛事。实时跟踪比赛，长时视频比赛和深度图像视频比赛分别于 2017 - 2019 年相继开展。关于热成像视频视频，VOT-15，16 和 17[114] 推出了仅热成像图像视频用于目标跟踪的数据集，之后 VOT-19 和 20[117] 新推出同步采集彩色和热成像视频的 RGB-T 数据集。

表 2: VOT 视频目标跟踪挑战赛统计

| | 短时视频 (VOT-ST) | 热成像视频 (VOT-T) | 实时跟踪 (VOT-RT) | 长时视频 (VOT-LT) | 颜色+深度视频 (VOT-RGBD) | 颜色+热成像视频 (VOT-RGBT) |
|---|---|---|---|---|---|---|
| VOT-2013[111] | ✓ | ✗ | ✗ | ✗ | ✗ | ✗ |
| VOT-2014[112] | ✓ | ✗ | ✗ | ✗ | ✗ | ✗ |
| VOT-2015[113] | ✓ | ✓ | ✗ | ✗ | ✗ | ✗ |
| VOT-2016[114] | ✓ | ✓ | ✗ | ✗ | ✗ | ✗ |
| VOT-2017[115] | ✓ | ✓ | ✓ | ✗ | ✗ | ✗ |
| VOT-2018[116] | ✓ | ✗ | ✓ | ✓ | ✗ | ✗ |
| VOT-2019[117] | ✓ | ✗ | ✓ | ✓ | ✓ | ✓ |
| VOT-2020[118] | ✓ | ✗ | ✓ | ✓ | ✓ | ✓ |

表 3：视频目标跟踪数据集算法结果评估

| 算法 | OTB-2015 | | VOT-2016 | | LaSOT | | 速度 | 出处 |
|---|---|---|---|---|---|---|---|---|
| | Prec.@20 | Succ.AUC | Accuracy | EAO | Precision | Succ.AUC | | |
| MOSSE[12] | 41.4 | 31.1 | - | - | - | - | 355fps | CVPR10 |
| KCF[27] | 69.5 | 47.7 | 0.49 | 0.192 | 16.6 | 17.8 | 212fps | TPAMI 15 |
| SAMF[30] | 75.3 | 55.2 | 0.51 | 0.186 | 20.3 | 23.3 | 17fps | ECCVW14 |
| DSST[42] | 69.3 | 47.0 | 0.53 | 0.181 | 18.9 | 20.7 | 28fps | CVPR14 |
| Staple[32] | 78.4 | 57.9 | 0.54 | 0.295 | 23.9 | 24.3 | 60fps | CVPR16 |
| HCF[35] | 83.7 | 56.2 | 0.44 | 0.220 | 25.6 | 25.1 | 11fps | ECCV16 |
| SRDCF[45] | 78.8 | 59.8 | 0.53 | 0.244 | 21.9 | 24.5 | 3fps | ICCV 15 |
| DeepSRDCF[120] | 85.1 | 63.5 | 0.51 | 0.276 | - | - | 1fps | ICCVW 15 |
| C-COT[37] | 89.6 | 66.7 | 0.54 | 0.331 | - | - | 1fps | ECCV16 |
| ECO[38] | 90.9 | 68.7 | 0.55 | 0.363 | 30.1 | 32.4 | 5fps | CVPR17 |
| ECO-HC[38] | 84.1 | 63.0 | 0.54 | 0.300 | 27.9 | 30.4 | 15fps | CVPR17 |
| UPDT[41] | 93.1 | 70.1 | - | - | - | - | - | ECCV18 |
| LMCF[85] | - | 56.8 | - | - | - | - | 80fps | CVPR17 |
| BACF[48] | 81.6 | 61.5 | 0.51 | 0.223 | 23.9 | 25.9 | 25fps | ICCV 17 |
| CSR-DCF[33] | 77.7 | 57.6 | 0.51 | 0.338 | 22.0 | 24.4 | 13fps | CVPR17 |
| STRCF[46] | - | 65.1 | 0.53 | 0.279 | 29.2 | 31.5 | 29fps | CVPR18 |
| DSAR-CF[25] | 83.2 | 63.9 | 0.54 | 0.258 | - | - | 6fps | TIP 19 |
| SSR-CF[87] | 84.5 | 64.5 | 0.53 | 0.248 | 26.6 | 27.4 | 23fps | TIP 19 |
| WSCF[50] | 82.6 | 62.4 | 0.55 | 0.283 | 24.4 | 26.5 | 16fps | TIP 20 |
| ASRCF[49] | 91.9 | 68.9 | 0.56 | 0.391 | 33.7 | 35.9 | 28fps | CVPR19 |
| SiameseFC[24] | 77.1 | 58.2 | 0.53 | 0.235 | 42.0 | 33.6 | 86fps | ECCVW16 |
| DSiam[65] | 81.5 | 60.5 | - | - | 40.5 | 33.3 | 45fps | ICCV 17 |
| SA-Siam[121] | 86.5 | 65.7 | 0.54 | 0.291 | - | - | 50fps | CVPR18 |
| RASNet[62] | - | 64.2 | - | - | - | - | 83fps | CVPR18 |
| SiamFC-tri[75] | 78.0 | 59.0 | - | - | - | - | 22fps | ECCV18 |
| SiameseRPN[53] | 85.1 | 63.7 | 0.56 | 0.344 | - | - | 200fps | CVPR18 |
| DaSiameRPN[55] | 88.1 | 65.8 | 0.61 | 0.411 | - | 41.5 | 160fps | ECCV18 |
| SiameseRPN++[60] | 91.5 | 69.6 | - | - | 56.9 | 49.6 | 35fps | CVPR19 |
| SPM[56] | 89.9 | 68.7 | 0.62 | 0.434 | - | - | 120fps | CVPR20 |
| SiamDW[61] | 85.0 | 64.0 | 0.54 | 0.300 | 47.6 | 38.5 | 67fps | CVPR19 |
| SiamAttn[63] | 92.6 | 71.2 | 0.68 | 0.537 | 64.8 | 56.0 | 45fps | CVPR20 |
| SiamR-CNN[88] | 89.1 | 70.1 | - | - | 72.2 | 64.8 | 5fps | CVPR20 |
| SiamCAR[71] | - | - | - | - | 60.0 | 50.7 | 52fps | CVPR20 |
| SiamBAN[72] | 91.0 | 69.6 | - | - | 59.8 | 51.4 | 40fps | CVPR20 |
| Ocean[73] | 92.0 | 68.4 | - | - | - | 56.0 | 58fps | ECCV20 |
| PG-Net[92] | 89.2 | 69.1 | - | - | 60.5 | 53.1 | 42fps | ECCV20 |
| CLNet[66] | - | - | - | - | 57.4 | 49.9 | 46fps | ECCV20 |
| MDNet[22] | 90.9 | 67.8 | 0.53 | 0.257 | 37.3 | 39.7 | 1fps | CVPR16 |
| TCNN[77] | - | - | 0.55 | 0.325 | - | - | 1fps | ECCVW16 |
| CREST[78] | 83.7 | 62.3 | 0.51 | 0.283 | - | - | 1fps | ICCV 17 |
| RFL[79] | 77.8 | 58.1 | 0.52 | 0.223 | - | - | 15fps | ICCVW 17 |
| MemTrack[68] | 82.0 | 62.6 | 0.53 | 0.273 | - | - | 50fps | ECCV18 |
| GCT[80] | 85.3 | 64.7 | - | - | - | - | 50fps | CVPR19 |
| ATOM[81] | 87.9 | 66.7 | - | - | 57.6 | 51.5 | 30fps | CVPR19 |
| DiMP[83] | 89.9 | 68.6 | - | - | 64.8 | 56.8 | 43fps | ICCV 19 |
| PrDiMP[84] | - | 69.6 | - | - | - | 59.8 | 30fps | CVPR20 |

### 4.2 数据集评估方法

表 3 展示了第三章主要介绍的视频目标跟踪算法在三个数据集上的评估结果，这里采用了三个数据集，分别是使用范围最广的经典数据集 OTB-2015，以及挑战赛数据集 VOT-2016，还使用了目前最大的长时视频高质量数据集 LaSOT†。在 OTB 数据集上，我们采用了经典的算法评估指标目标中心定位精度(Prec.@20) 以及目标跟踪正确率(Succ.AUC)。前者是指算法预测的目标中心位置与真实结果的误差，这里以 20 像素为阈值，定位精度误差小于 20 像素则认为跟踪结果正确。 后者更加综合地反映了目标跟踪的正确率，该指标以预测目标的矩形框与真实结果的交并比为衡量依据，绘制不同交并比阈值下的正确率曲线，并计算曲线的下方图形面积 (Area Under the Curve, AUC) 作为结果。与 OTB 不同，VOT 数据集采取了新的算法评估方式。 VOT 引入了重新初始化的机制，即当跟踪算法对目标跟踪结果判定为失败时，重新给定当前目标的正确位置继续跟踪，并记录重新初始化次数，即丢失目标次数用于评估算法的鲁棒性。VOT 上的精度 (Accuracy) 反映目标跟踪结果与真实结果的定位误差，而综合指标 EAO (Expected Average Overlap) 反映了算法的目标定位精度以及鲁棒性。需要说明的是，最新的 VOT 2020 取消了重启机制，转而使用初始化点 (initialization points) 代替。即在每个序列中，在初始帧、结束帧、以及期间每隔固定的帧，都设置一个初始化点，参评的跟踪器将从初始化点开始运行，从而避免了重启机制对跟踪器评估不够鲁棒和不够公平的问题。LaSOT 采用了与 OTB-2015 相同的评价指标。

### 4.3 算法结果及分析

表 3 第一部分展示了相关滤波目标跟踪算法的评估结果以及算法速度的比较。早期的 MOSSE 算法尽管精度现在看来比较低，但可以达到超高的 355 fps 的 CPU 运行速度，为后续其它算法对其进行改进提奠定良好的基础。KCF 是将相关滤波目标跟踪算法推向主流的关键工作，其在 MOSSE 的基础上通过引入多通道 HOG 特征，很大程度地提升了算法的精度，在 OTB-2015 数数据集上跟踪定位精度提升 28%，跟踪正确率提升 16%，并仍保持超高的运行速度。后续出现的两类经典的尺度估计方法 SAMF 和 DSST 考虑了目标尺度的变化，虽然没有很大程度地提升算法精度，但为后续的算法实现目标变化尺度估计奠定了良好的基础。由于引入了多尺度重复检测的机制，因此算法速度上有较大程度的损失。之后 Staple 和 HCF 算法分别引入了颜色特征以及深度特征加强目标的表征能力，进一步提升了目标跟踪的精度。同时可以看到，深度特征的引入，也使得 HCF 的算法速度进一步降低至 11 fps。与此同时，解决相关滤波目标跟踪中的边界效应问题受到了研究者的关注。SRDCF 首先引入了空间正则化项，很大程度上缓解了相关滤波目标跟踪的边界效应问题，使得算法仅使用 HOG 特征，就达到了与 Staple 以及 HCF 接近的跟踪精度。而同时加入深度特征的 DeepSRDCF 算法更是达到了当时最先进的跟踪精度。尽管算法精度得到了明显的提升，但是可以看到 DeepSRDCF 算法的运行速度仅有 1 fps，相较于原始的 MOSSE 以及 KCF 算法速度已经下降百余倍，这使得 DeepSRDCF 与目标跟踪算法满足实时性的基本需求也有很大的差距。在 SRDCF 的基础上，一些方法如 STRCF，DSAR-CF，SSR-CF，ASRCF 从不同的角度对 SRDCF 进行优化，不仅在算法精度上得到了提升，算法速度也得到了不同程度的改善。最新的 ASRCF 算法已经可以在多个数据集上达到先进的跟踪精度并保持接近实时的运行速度。另外，BACF 以及 WSCF 从样本生成的角度出发解决相关滤波目标跟踪的边界效应问题，也取得了相较于 SRDCF 更优良的跟踪精度及运行速度。还有同期的 LMCF 和 CSR-DCF 从目标跟踪任务所面临的挑战出发，提出了针对性的应对策略，也取得了较好的效果。最后，C-COT 从新的角度出发实现了特征空间、相关滤波、响应图的连续化，大大提升了算法的精度，不过大量的插值操作极大地影响了算法速度，使得算法效率极低，仅有小于 1 fps 的速度。在此基础上，ECO 通过多项加速策略一定程度上改善了算法的效率问题，同时提出的 ECO-HC 放弃了深度特征，采用 HOG 特征和颜色特征，进一步提升了算法的速度。 最后，UPDT 算法通过发掘深度特征的力量，再次刷新了 OTB-2015 数据集上的跟踪精度，目前为止仍保持该数据集上的最优结果。

† 部分算法由于与 LaSOT 数据集发表时间接近，因此未参与该数据集上的算法评估。

表 3 第二部分展示了基于孪生网络 (Siamese Network) 的目标跟踪算法的评估结果以及算法速度的比较。早期的 SiameseFC 算法具备良好的跟踪精度并保持较高的运行速度，在 GPU 上保持 86 fps。在此基础上的改进方法，如 DSiam, SA-Siam, RASNet, SiamFC-tri, SiamDW 算法在跟踪精度上相较于 SiameseFC 有不同程度的提升，同时也带来了一定的速度损耗。另外，SiameseRPN[53] 算法从新的角度出发将目标检测的目标候选框思想引入相关滤波目标跟踪算法，相较于 SiameseFC 在算法的跟踪精度和运行速度方面均有了明显的提升。与其他算法框架类似，在 SiameseRPN 的基础上，DaSiamRPN[55] 以及 SiameseRPN++[60] 算法进一步提升了跟踪的精度，也一定程度上影响了运行速度，不过值得一提的是，此类算法均保持着超过实时的 (GPU) 运行速度。最新的算法，如 Siam R-CNN[88] 在 LaSOT 数据集上达到了领先的精度，但是在速度上有所缺陷，仅达到约 5 fps 的运行速度。一些无需锚点先验的方法，如 SiamCAR[71]，SiamBAN[72]，Ocean[73] 在跟踪精度上超过了 SiameseRPN 方法，且保持了较高的 50+ fps 左右的算法速度。

表 3 第三部分展示了基于其它深度网络的目标跟踪算法的评估结果以及速度。可以看到，早期的 MDNet, TCNN 算法在算法精度上具有明显的优势，不仅分别取得 VOT-2015 和 VOT-2016 挑战赛的冠军，而且其算法精度与之后几年提出的算法相比仍具有一定的优势。然而，此类算法最大的缺点是运行速度的限制，这也是早期基于深度学习的目标跟踪方法的主要问题所在。之后出现的基于循环卷积网络 RNN 的算法 RFL 和 MeemTrack 虽然在精度上并不特别突出，但是兼顾了算法速度。最近的 GCT 将图卷积网络 GCN 引入目标跟踪，也达到了先进的跟踪精度和实时的 (GPU) 运行速度。值得一提的是，ATOM[81] 及其系列算法 DiMP[83]，PrDiMP[84] 达到了当时最先进的跟踪精度，同时维持着较高的 30+ fps 的实时运行速度。

表 4 展示了视频目标跟踪挑战赛 VOT 自 2013 年至 2020 年的比赛排名前十名的算法及结果评估†。可以看到，在 2013 年参赛的目标跟踪算法，如 PLT 等，主要基于 SVM 分类器，稀疏表示等经典机器学习方法。而在 2014 年，相关滤波目标跟踪算法出现并迅速表现出强大的性能，基于相关滤波的跟踪算法 DSST, SAMF, KCF 包揽了 VOT-2014 比赛的前三名，超过了 VOT-2013 的冠军方法 PLT。从此，相关滤波目标跟踪算法长期在 VOT 比赛上占据了主导地位。VOT-2015 上，深度学习方法开始崭露头角，MDNet 凭借深度网络的强大学习能力，在精度上超过了当年最好的相关滤波目标跟踪算法 DeepSRDCF ，获得当年比赛的冠军。基于相关滤波的算法如 DeepSRDCF, SRDCF, NSAMF 也具有良好的表现。到 VOT-2016，排名第一的方法是结合深度特征与相关滤波算法的 C-COT，也是目前为止唯一一个蝉联两届 VOT 比赛第一名的算法。与此同时，基于深度卷积网络 CNN 的算法 TCNN，也取得了第二名的成绩。到 2017 年，除了排名第一的 C-COT 外，其他排名前五的方法，包括 CFCF, ECO, Gnet, CFWCR 也均为基于相关滤波的目标跟踪算法。同样，VOT-2018 上，排名前五的算法，包括 MFT，UPDT，RCO，LADCF，DeepSTRCF 仍然全部为基于相关滤波的目标跟踪算法。 由此可见基于相关滤波的目标跟踪算法在视频目标跟踪任务上展现了强大的优势。直到 2019 年，基于深度网络的目标跟踪算法才在目标跟踪比赛上展现出强大实力，如基于卷积神经网络的 ATP 取得了 VOT- 2019 的第一名。而第二名的方法 DiMP 利用相关滤波算法在线学习的思想结合深度卷积网络结构进行融合而得到。在 VOT-2020 上，冠军方案 RPT 利用目标状态估计网络与在线分类网络，提升了对目标位姿变化、几何结构变化的建模能力。其他基于深度网络的目标方法几乎包揽了竞赛的前列。

表 5 展示了视频目标跟踪挑战赛实时算法的排名及结果。VOT-RT2017 挑战赛的第一名是基于相关滤波的目标跟踪算法 CSRDCF[33] 的改进版本，在具备实时性的目标跟踪算法中，CSRDCF++ 展现了绝对的领先的精度优势。同时可以看到，基于孪生网络的 SiamFC 算法取得了当年比赛的第二名。而到 2018 年 VOT 实时比赛，基于孪生网络的目标跟踪算法展现了绝对了优势，前十名的算法中，基于孪生网络的方法占了八个，包括 SiamRPN, SA_Siam_R, SA_Siam_P, SiamVGG, LWDNTm, LWDNTthi, MBSiam, UpdateNet 均由 SiameseFC[24] 算法衍生而来。再到 VOT-RT2019，

† VOT 2018 和 2019 大赛组委会只公开了内测排名前八 和前五的算法及结果。

表 4: VOT 视频目标跟踪挑战赛结果评估

| | VOT-13 | | VOT-14 | | VOT-15 | | VOT-16 | |
|---|---|---|---|---|---|---|---|---|
| 排名 | 算法 | $R_\Sigma \downarrow$ | 算法 | $R_\Sigma \downarrow$ | 算法 | EAO ↑ | 算法 | EAO ↑ |
| No.1 | PLT | 4.48 | DSST[42] | 8.77 | MDNet[22] | 0.38 | C-COT[37] | 0.331 |
| No.2 | FoT | 7.33 | SAMF[30] | 9.10 | DeepSRDCF[120] | 0.32 | TCNN[77] | 0.325 |
| No.3 | EDFT | 9.85 | KCF[27] | 9.33 | EBT | 0.31 | SSAT | 0.321 |
| No.4 | LGT++ | 10.05 | DGT | 9.48 | SRDCF[45] | 0.29 | MLDF | 0.311 |
| No.5 | LT-FLO | 11.03 | PLT_14 | 9.51 | LDP | 0.28 | Staple[32] | 0.295 |
| No.6 | GSDT | 11.07 | PLT_13 | 10.62 | sPST | 0.28 | DDC | 0.293 |
| No.7 | SCTT | 11.29 | sASMS | 12.85 | SC-EBT | 0.25 | EBT | 0.291 |
| No.8 | CCMS | 11.36 | HMM-TxD | 14.33 | NSAMF[30] | 0.25 | SRBT | 0.290 |
| No.9 | LGT | 11.61 | MCT | 14.61 | Struct | 0.25 | STAPLE+[32] | 0.286 |
| No.10 | Matrioska | 11.68 | MatFlow | 15.51 | RAJSSC | 0.24 | DNT | 0.278 |
| | **VOT-17** | | **VOT-18** | | **VOT-19** | | **VOT-20** | |
| 排名 | 算法 | EAO ↑ | 算法 | EAO ↑ | 算法 | EAO ↑ | 算法 | EAO ↑ |
| No.1 | C-COT[37] | 0.203 | MFT | 0.2518 | ATP | 0.2747 | RPT | 0.530 |
| No.2 | CFCF[39] | 0.202 | UPDT[41] | 0.2469 | DiMP[83] | 0.2489 | OceanPlus[73] | 0.491 |
| No.3 | ECO[38] | 0.196 | RCO | 0.2457 | DRNet | 0.2371 | AlphaRef | 0.482 |
| No.4 | Gnet | 0.196 | LADCF | 0.2218 | Cola | 0.2218 | AFOD | 0.472 |
| No.5 | CFWCR[122] | 0.187 | DeepSTRCF[46] | 0.2205 | Trackyou | 0.2035 | LWTL | 0.463 |
| No.6 | LSART | 0.185 | CPT | 0.2087 | | | fastOcean[73] | 0.461 |
| No.7 | MCCT | 0.179 | SiamRPN[53] | 0.2054 | | | DET50 | 0.441 |
| No.8 | MCPF | 0.165 | $\text{DLST}_{\text{pp}}$ | 0.1961 | | | D3S[90] | 0.439 |
| No.9 | SiamDCF | 0.160 | | | | | Ocean[73] | 0.430 |
| No.10 | CSRDCF[33] | 0.150 | | | | | TRASTmask | 0.370 |

表 5：VOT 视频目标跟踪挑战赛实时算法结果评估

| | VOT-RT17 | | VOT-RT18 | | VOT-RT19 | | VOT-RT20 | |
|---|---|---|---|---|---|---|---|---|
| 排名 | 算法 | EAO ↑ | 算法 | EAO ↑ | 算法 | EAO | 算法 | EAO ↑ |
| No.1 | CSRDCF++[33] | 0.212 | SiamRPN[53] | 0.383 | SiamMargin | 0.366 | AlphaRef | 0.486 |
| No.2 | SiamFC[24] | 0.182 | SA_Siam_R | 0.337 | SiamFCOT | 0.350 | OceanPlus[73] | 0.471 |
| No.3 | ECO-HC[38] | 0.177 | SA_Siam_P[121] | 0.286 | DiMP[83] | 0.321 | AFOD | 0.458 |
| No.4 | Staple[32] | 0.170 | SiamVGG | 0.275 | DCFST | 0.317 | fastOcean[73] | 0.452 |
| No.5 | KFebT | 0.169 | CSRTPP | 0.263 | SiamDW-ST[61] | 0.299 | Ocean[73] | 0.419 |
| No.6 | ASMS | 0.168 | $\text{LWDNT}_{\text{thi}}$ | 0.262 | ARTCS | 0.287 | D3S[90] | 0.416 |
| No.7 | SSKCF | 0.164 | $\text{LWDNT}_{\text{m}}$ | 0.261 | SiamMask[89] | 0.287 | AFAT | 0.372 |
| No.8 | CSRDCFf | 0.158 | CSTEM | 0.239 | SiamRPN++[60] | 0.285 | SiamMargin | 0.355 |
| No.9 | UCT | 0.145 | MBSiam | 0.238 | SPM[56] | 0.275 | LWTL | 0.337 |
| No.10 | MOSSE_CA[91] | 0.139 | UpdateNet | 0.209 | SiamCRF-RT | 0.262 | TRASTmask | 0.321 |

前十名的算法中也有七个基于孪生网络算框架，包括 SiamMargin, SiamFCOT, SiamDWST, SiamMask, SiamRPN++, SPM 和 SiamCRF-RT。VOT-RT2020 上，基于无锚点 Siamese 网络的算法 Ocean[73] 及其衍生算法 OceanPlus，fastOcean 就包揽了前五名中的三项。由此可见，兼具精度和速度的孪生网络的目标跟踪算法在目标实时跟踪任务上展现了巨大的优势。

### 4.4 目标跟踪算法实验分析小结

通过上述对当前主流的视频目标跟踪算法在各个数据集和历年挑战赛上的结果比较和分析，我们可以梳理出视频目标跟踪近十年来的发展脉络：

1) 算法追求精度与速度兼顾：从表 3 可以看出，最早期的相关滤波目标跟踪算法可以达到超过 300 fps 的运行速度，然后随着特征的加强与模型复杂度的增加，虽然跟踪精度得到一步步提升，但算法的速度也大幅度下降，发展到 C-COT 算法，虽然蝉联了两届 VOT 挑战赛的第一名，但不到 1 fps 的跟踪速度使得算法难以实际应用。因此后续相关滤波算法开始进一步对冗余的模型参数，特征等进行优化，使得算法接近实时的运行速度。类似的，基于孪生滤波的算法早期也具有较高的运行速度，如 SiameseFC 可以达到 86 fps，之后的改进算法在牺牲一定速度的前提下提升了精度，但都保持了超过或接近实时的运行速度。因此，兼顾算法精度与速度是最近年来目标跟踪算法的重要发展趋势之一。

2) 相关滤波和孪生网络算法框架优势明显：从各个数据集上算法评估以及 VOT 比赛结果来看，相关滤波和孪生网络算法框架已发展成为近年来目标跟踪问题的主流算法框架。

3) 相关滤波算法精度优势突出：近年来，相关滤波目标跟踪算法在视频目标跟踪领域各大数据集上表现良好，具备强大的性能和明显的优势。回顾 VOT-2013 至 VOT-2020 视频目标跟踪比赛，如表 4 所示，相关滤波目标跟踪算法在参赛数量和结果表现上均处于领先地位。然而，伴随着近年来的快速发展，相关滤波目标跟踪算法也已经发展得比较成熟，算法改进和提升的空间也相对有限。

4) 孪生网络算法综合性能良好：近年来基于孪生网络的视频目标跟踪算法由于兼顾实时性与精度，综合性能表现良好。从 VOT-RT2017 至 VOT-RT2020 视频目标跟踪实时算法比赛结果来看，基于孪生网络的算法表现突出。

5) 深度学习方法的作用日益突出：不论是相关滤波算法融合深度特征对效果有明显提升，还是其他深度网络的目标跟踪算法快速发展，都说明了深度学习方法在视频目标跟踪任务中的作用日益显现。

## 5 视频目标跟踪发展趋势

本章将主要介绍讨论视频目标跟踪的未来发展趋势和和研究建议。首先，针对目标跟踪发展面临的痛点，如目前算法无法适用于长时间视频，低功耗设备、复杂对抗环境等，5.1 节重点介绍了目标跟踪算法距离实际落地应用面临技术瓶颈。为实现更加鲁棒的跟踪，考虑增加信息源，如多模态信息，也是提升目标跟踪鲁棒性的重要手段之一，5.2 节重点介绍了多模态数据下的目标跟踪相关研究。最后，为探究目标跟踪更多的应用情形与交叉研究，5.3 节介绍了目标跟踪与计算机视觉领域其他密切相关任务，如视频目标检测、目标分割的交叉研究。

### 5.1 目标跟踪发展瓶颈与解决方案

#### 5.1.1 长时间视频目标跟踪研究

目前的视频目标跟踪算法与现实应用还存在较大的差距，其中最重要的问题之一就是算法对目标的长时间跟踪。现有的目标跟踪算法往往在平均时长为 1,000 帧以内的视频上进行测试和评估，即便是最近提出的数据集，如 LaSOT[17]，TrackingNet[106] 等，包含的长时视频的平均长度也往往保持在 2-3 分钟，这距离实际生活应用中的视频差距较大。长时视频目标跟踪的困难在于，很多跟踪器在目标发生丢失之后，很难重新对目标完成重跟踪。对于长时间跟踪，跟踪器一旦丢失目标后，后续的视频序列往往变为无效跟踪。因此，长时间目标跟踪更具有实用性，从最新的数据集和挑战赛的发展也可以看出长时间目标跟踪受到了越来越多研究者的关注和重视，是视频目标跟踪领域未来重要的发展方向之一。早期的工作[123] 就开始关注长时间目标跟踪问题，其基于相关滤波算法试图解决长时间目标跟踪中可能面临的目标遮挡、变形、出视野等问题。最近的 SPLT[124] 利用相同的 SiamPRN[53] 算法进行跟踪和重检测，保证目标长时间丢失后可以重新跟踪，并利用其提出的略读模型对算法进行加速。此外，最新方法 LTMU[125] 专门针对长时间目标跟踪问题，同时利用了在线局部

跟踪器以及验证器，并提出了中间更新器对模型是否更新进行判定，用于长时间目标跟踪，并取得了显著的效果。相关工作也获得了 CVPR 2020 最佳论文提名。尽管如此，长时间目标跟踪仍然具备巨大的挑战性，需要进一步研究和探索。另外，结合视频目标跟踪与其他研究问题，如目标重检测、行人重识别等，解决目标丢失后如何实现目标重新跟踪，将会在一定程度上为长时目标跟踪提供帮助。

5.1.2 低功耗设备目标跟踪研究

如何实现算法精度与速度的平衡是视频目标跟踪问题一直以来面临的难题。作为视频目标跟踪的主流方法之一，相关滤波目标跟踪算法在开创之初，具有普通 CPU 设备上远超实时的跟踪速度。然而，随着所采用的特征的多样，尤其是深度特征的加入，以及模型复杂度的提升，最新的相关滤波目标跟踪经典算法，如 ECO[38]，也仅仅在 GPU 上达到接近实时性能的运行速度，在 CPU 上的运行速度远低于实时性。视频目标跟踪的另一主流方法孪生网络目标跟踪算法，作为基于深度网络的方法，为充分发挥算法速度优势，也需要高性能 GPU 设备作为支撑。而视频目标跟踪的实际应用场景往往是在车载相机车辆跟踪（无人驾驶技术），监控相机行人管控（智能监控技术）等，因此在低功耗设备上实现具备实时性能的目标跟踪是非常具有研究价值的方向，目前也几乎没有专门针对此类型问题的研究工作。

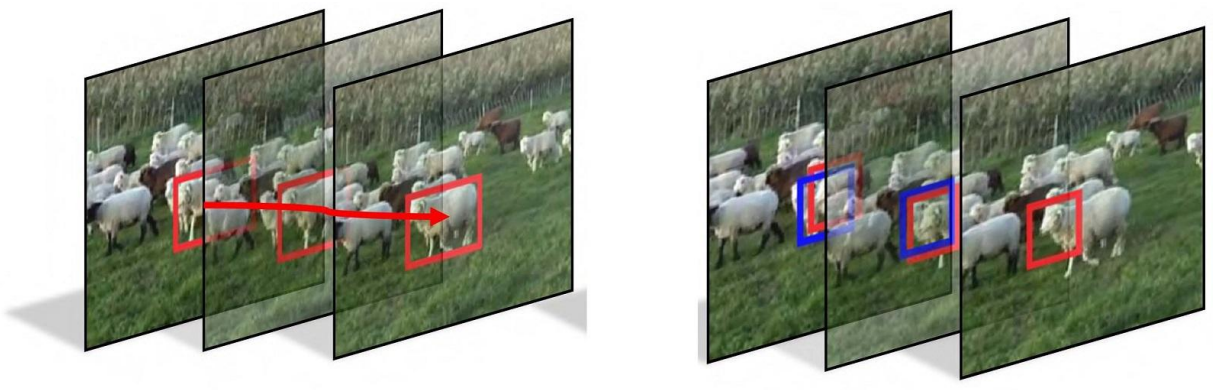

图 18. UDT 算法实现无监督深度网络训练

5.1.3 无标注训练目标跟踪研究

基于深度学习的目标跟踪算法往往需要大量的训练数据支撑，实验也表明更大量的训练数据可以使目标跟踪算法获得更高的精度，如 SiamRPN 算法在大规模数据库 Youtube-BB 上进行训练后跟踪结果获得明显提升。然而，对大规模训练数据的依赖也同时造成了深度目标跟踪算法训练成本高，时间长。近年来，随着半监督、无监督学习方法的问世和普及，越来越多的领域开始考虑利用更少的样本完成模型训练任务，同时保持算法精度。Wang 等人[126] 首先提出了无监督学习的深度跟踪 UDT (Unsupervised Deep Tracking) 算法，如图 18 所示，UDT 算法利用目标前向跟踪预测（红色框及箭头方向）与结果反向跟踪（蓝色框）的响应图之间的一致性计算网络损失，实现了网络的前向和反向传播。该方法首次将无监督学习的思想引入深度目标跟踪算法，实现了无标注样本集上的训练。

5.1.4 攻击鲁棒性目标跟踪研究

算法安全和鲁棒性一直是计算机领域关注的热点问题，近年来对抗攻击的研究开始受到关注，其可以为设计更加鲁棒的算法提供思路。目前，对抗攻击在目标跟踪领域尚未引起足够的重视。在文献[127]中，作者针对目前最新的的 SiamRPN++[60] 跟踪算法设计了名为“Cooling- Shrinking Attack”的对抗扰动生成器，实验结果表明，方法能够使 SiamRPN++ 算法的跟踪精度明显下降。几乎同期的几个工作，如 Spark (Spatial- aware Online Incremental Attack)[128]，FAN (Fast At- tack Network)[129]，RTAA (Robust Tracking against Adversarial Attacks)[130] 也各自从不同角度，设计了不同的攻击手段，用于研究对抗攻击对目标跟踪算法产生的影响与威胁。今后，面向日益重要的算法安全问题，对视频目标跟踪的对抗攻击与防御算法研究也将成为不可或缺的关键一环。

5.1.5 特定场景目标跟踪研究

通用场景下是视频目标跟踪是当前研究的热点，同时，针对特殊场景下的目标跟踪也越来越引起一些研究学者的关注。例如无人机航拍视频目标跟踪[131, 132]，遥感图像视频目标跟踪[133]。关于无人机航拍视频，基于相关滤波目标跟踪框架，最新的工作 ARCF[134] 通过判别并删除异常轨迹提高跟踪算法的性能，AotuTrack 则提出基于时空自适应空间正则化方法的无人机视频目标跟踪方法[135]。对于卫星视频目标跟踪, Shao 等人利用利用速度特征[136]，以及光流与方向梯度直方图组合特征[137]，基于相关滤波目标跟踪框架，实现更鲁棒的遥感图像目标跟踪。今后，开发更多特定场景下的目标跟踪，如水下场景的物体跟踪，医疗影像中的细胞跟踪，夜视图像目标跟踪等，将是拓展目标跟踪研究范畴与应用场景的重要发展方向之一。

### 5.2 多模态目标跟踪

5.2.1 基于 RGB-D 视频的目标跟踪

由于普通 RGB 相机的广泛应用，现有的目标跟踪算法和数据集大多关注于 RGB 视频。随着近年来 RGB-D 相机，激光雷达的普及，以及时间飞

行算法 (ToF) 等技术的成熟，一些研究者开始关注基于 RGB 图像和深度图像结合的视频目标跟踪算法。如图 19 所示，相比于颜色信息，深度信息可以有效地帮助视频实现前景背景分离，同时为目标遮挡判定提供有效的指导，这对解决视频目标跟踪的一些主要难题在信息源上提供了有力的保障。早在文献[138]中，作者就提出了基于 RGB-D 数据的目标跟踪数据集和算法，最近的 CDTB[108] 构建了更大规模的数据集用于算法评估。近年来，一些面向 RGB-D 视频数据的跟踪算法相继问[139 - 141]，不过相对于基于 RGB 视频的算法快速发展并取得显著成效，基于 RGB-D 视频的目标跟踪仍然发展缓慢，是未来值得进一步研究的方向之一。

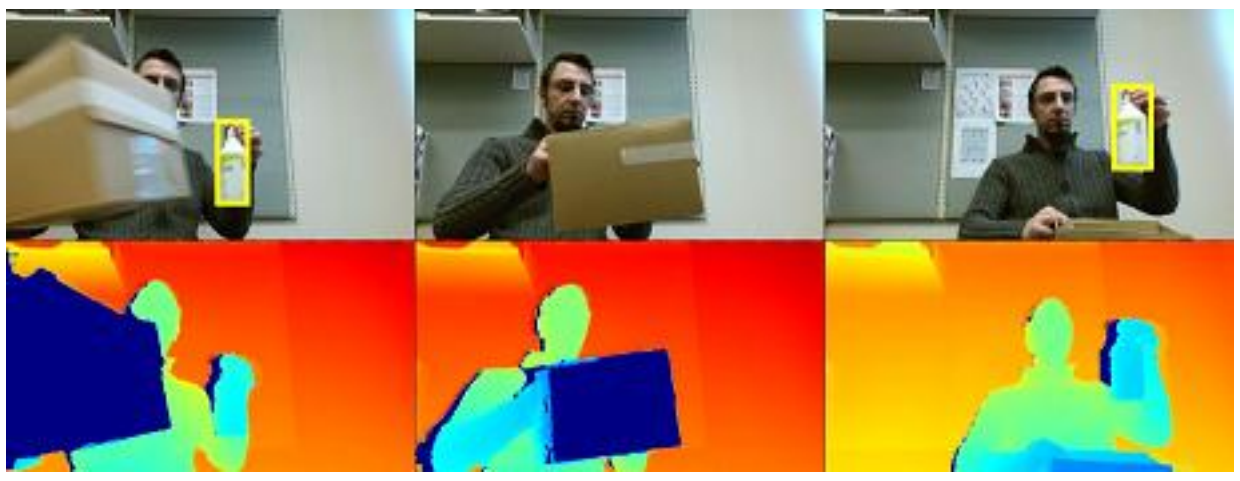

图 19. RGB-D 目标跟踪数据集 CDTB[108]

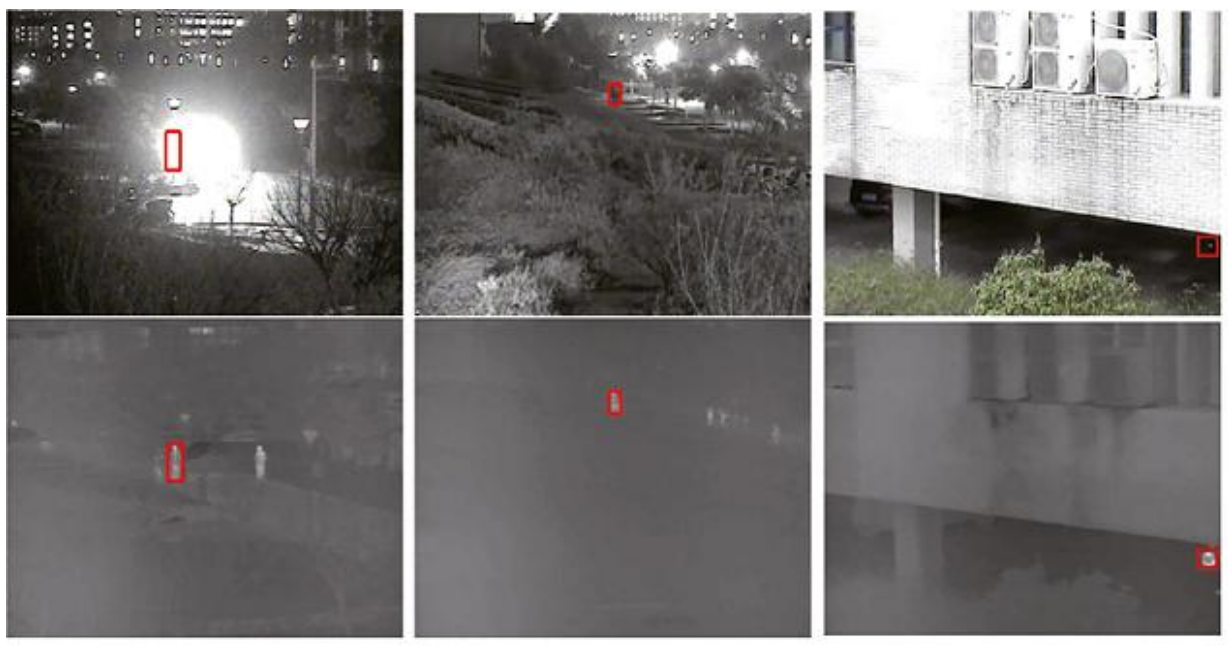

图 20. RGB-T 目标跟踪数据集 RGBT234[109]

5.2.2 基于 RGB-T 视频的目标跟踪

基于 RGB-T 视频的目标跟踪虽然具有一些研究工作，但相较于 RGB 视频，发展还相对欠缺。红外图像和 RGB 图像对于视频目标跟踪来说可以提供互补的信息，例如 RGB 图像可以提供丰富的目标外观颜色纹理信息，但在光照变化，雨雾场景下受到较大的影响，而红外图像往往不受此类情形的影响，如图 20 所示。与深度图像视频相似，红外图像信息往往也能帮助指导视频中目标的前背景分离。这是因为前景目标和背景通常具有不同的热力值。文献[142] 首次利用基于深度学习的方法解决 RGB-T 目标跟踪问题，接下来的工作[143 - 146] 先后探究了如何更好地结合不同模态下的数据，最新的研究[147] 从目标跟踪面临的各种挑战入手提升 RGB-T 目标跟踪的效果。今后，随着数据采集硬件设备的升级，利用多元互补信息实现更高精度更鲁棒的视频跟踪也是未来的重要研究方向之一。

### 5.3 目标跟踪交叉领域研究

5.3.1 目标检测与目标跟踪

视频目标跟踪与目标检测有密切的联系，但也有着本质的区别，主要表现在：(i) 目标检测面向特定类别的目标，而目标跟踪的对象是非特定目标；(ii) 目标检测不对同类别物体进行区分，但是目标跟踪对同类别物体区分十分重要；(iii) 考虑视频时序信息是目标跟踪的关键问题，但是在图像目标检测中并不涉及。为建立目标检测与目标跟踪之间的关联，TGM[148] 算法提出了目标指导模型 (Target-Guidance Module) 指导目标检测器定位跟踪目标相关的物体，并利用元学习的方法学习和建立了目标-干扰物在线分类模型，实现了基于任意目标检测器的目标跟踪统一框架。Siam R-CNN[88] 提出了基于孪生网络的目标重检测算法框架，进一步挖掘了两阶段目标检测算法在目标跟踪问题的潜力，通过将上述框架与动态规划算法结合，Siam R-CNN 充分考虑了视频第一帧给定目标模板与前一帧预测目标结果，使得算法可以对目标及背景干扰进行长时间的有效建模，在目标长时间丢失后实现重新检测继续跟踪，算法在目标跟踪任务，尤其是长时间目标跟踪任务上效果表现十分优良。Wang 等人[149] 将目标跟踪问题考虑为特定目标的检测问题，并将其称作实例级目标检测 (Instance Detection)。作者利用未知模型元学习方法 MAML (Model-Agnostic Meta-Learning) ，通过适当的模型初始化及对首帧图像待跟踪目标的学习，将目标检测器快速转化为目标跟踪器完成目标跟踪任务。

5.3.2 目标分割与目标跟踪

视频目标跟踪与视频目标分割也关联密切，前者仅要求用矩形（或多边形）框对目标进行定位和标注，而后者需要根据目标的形状轮廓完成准确的分割。视频目标跟踪可以为分割提供目标的初定位和分割初始化结果，而视频目标分割又可以帮助跟踪更好地完成前背景分离，从而学习目标前景特征[89, 90, 150, 151]。如图 21 所示，Wang 等人提出的 SiamMask 算法[89]，基于半监督学习思想，利用统一的框架同时完成视频目标跟踪和视频目标分割任务。D3S[90] 算法则应用具有几何互补信息双分支模型分别处理非刚性物体形变和刚性物体假设，构建单阶段 (one-stage) 模型，实现鲁棒的视频目标

在线分割和跟踪。可以看到，目标分割结果可以更精确地表征目标前景，因此在很多场景更加实用。

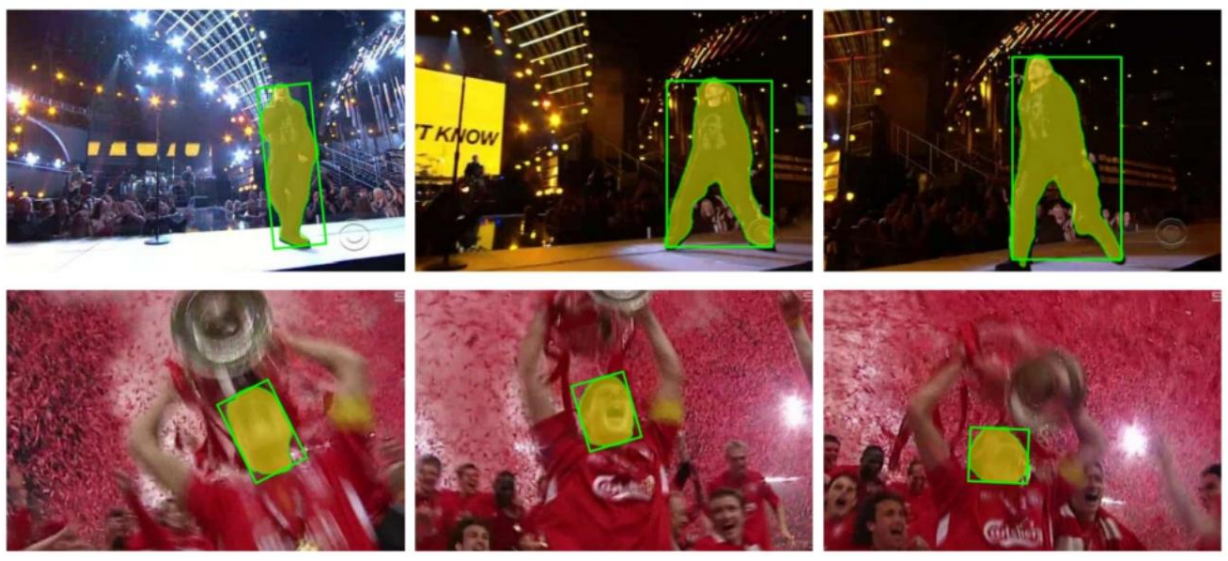

图 21. SiamMask 同时实现视频目标跟踪与分割

# 6 结论

本文就计算机视觉中的重要任务视频目标跟踪为主题展开综述。首先分类介绍了视频目标跟踪任务目前面临的诸多挑战，包括目标消失、目标变化、背景干扰、目标移动四类主要情形。接下来，本文详细介绍了近十年来目标跟踪领域的相关算法，包括该领域两大主流算法框架的基本原理以及经典方法和代表性工作，还介绍了其他基于深度学习的目标跟踪算法，以及介绍了解决目标跟踪面临挑战的典型应对策略。本文还详细介绍和比较了目标跟踪任务的公开数据集和挑战赛，总结了视频目标跟踪的历史发展脉络和未来发展方向。本文还概述了视频目标跟踪未来的发展趋势。针对如何解决目标跟踪任务面临的痛点和瓶颈，从长时间、低功耗、抗攻击、无监督目标跟踪算法和特定场景目标跟踪任务展开分析。此外，融合多模态数据，如深度图像、红外图像将会为视频目标跟踪带来更多新的研究问题和解决方案。最后，本文讨论了目标跟踪与视频目标检测，视频目标分割任务的关联与交叉研究。

**致　谢**　本课题得到天津市自然科学基金项目 No. 18JCYBJC15200，天津市研究生科研创新项目 No. 2021YJSB174，国家自然科学基金项目 (NSFC) No. U1803264 资助。本文的完成也感谢同实验室老师王松教授，万亮教授的指导和帮助，以及实验室成员张鹏宇，陈峙灏同学的日常学术研讨。

## 参 考 文 献

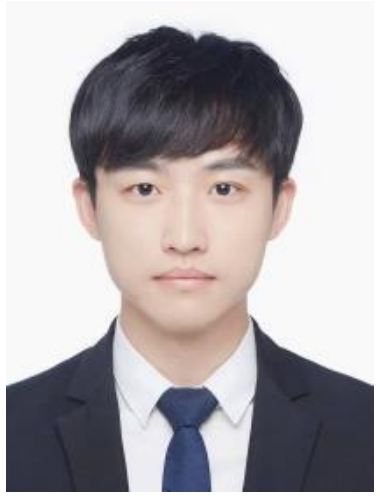

**HAN Rui-Ze**, born in 1994, Ph.D. candidate. His major research interests include multi-camera video analysis and visual object tracking. He is a student member of the CCF.

**FENG Wei**, born in 1978, Ph.D., Professor. His major research interests include active robotic vision and visual intelligence. He is a member of the CCF.

**GUO Qing**, born in 1989, Ph.D. His research interests include computer vision and visual object tracking.

**HU Qing-Hua**, born in 1976, Ph.D., Professor. His current research is focused on uncertainty modeling in big data, machine learning with multi-modality data. He is a member of the CCF.

**Background**

Visual object tracking is an important and fundamental task in computer vision and artificial intelligence, which has many real-world applications, e.g., video surveillance, visual navigation, etc. Visual object tracking also has many challenges, including object loss, object deformation, background clutters, and object fast motion, etc. For the accurate and efficient object tracking under the above scenario, many visual object tracking methods have been proposed in recent years. Among them, many machine learning technologies, e.g., support vector machines (SVM), subspace learning, sparse and compressive reconstruction, correlation filter (CF), convolutional neural network (CNN), recurrent neural network (RNN) and Siamese network, have been applied in the methods for visual tracking task.

This paper provides a survey on the research development of the visual object tracking. In this paper, we first detailedly state and illustrate the challenges in visual object tracking. Then, we present a comprehensive review of the rationale and technologies of the representative works on visual object tracking in recent ten years, we especially present the two most popular frameworks, i.e., the correlation filter (CF) and Siamese network, for visual tracking. This paper also systematically compares the data statistics of the common benchmarks for visual tracking and the evaluation results of the representative tracking algorithms. We summarize the development history of the visual tracking. We further infer the development trend of the visual object tracking in the future from there directions, including technical method research, practical application research and data collection source.

This research group has some achievement on the visual object tracking in the past few years. More than ten papers on this topic have been published on the top journal or conference in recent five years, including IEEE International Conference on Computer Vision (ICCV), Annual Conference on Neural Information Processing Systems (NeurIPS), IEEE Trans. On Image Processing (TIP). HAN Rui-Ze et al. won the Best Paper Award of IEEE International Conference on Multimedia and Expo (ICME) 2018, which focuses on the research of correlation filter based visual object tracking.

This work was supported by the Natural Science Foundation of Tianjin under Grant 18JCYBJC15200, Tianjin Research Innovation Project for Postgraduate Students under Grant 2021YJSB174, and the National Natural Science Foundation of China (NSFC) under Grant U1803264.